\documentclass[11pt,a4paper,logo,copyright]{googledeepmind}
\usepackage{gdm-colors}
\usepackage{flafter}

\title{Advancing Model Research in AgentX:\\Long-Horizon Autonomy for\\Industrial Recommender Systems}

\author{
Shuang~Yang\textsuperscript{*}, Zijie~Zhuang\textsuperscript{*,\textdagger}, Changxin~Lao\textsuperscript{*}, Pengbo~Xu\textsuperscript{*}, Hanwen~Xu\textsuperscript{*},
Yusheng~Huang, Han~Gao, Guanchen~Wang, Tianbao~Ma, Linxun~Chen,
Peilin~Song, Xuming~Wang, Chen~Li, Fan~Wu, Tao~Wang,
Zibo~Zhao, Xiangyu~Wu, An~Liu, Fei~Pan, Peng~Jiang, Chen~Yang,
Zhaojie~Liu, Wenwu~Ou}
\renewcommand{\today}{}
\newcolumntype{Y}{>{\raggedright\arraybackslash}X}

\fancypagestyle{agentxfirst}[firststyle]{
  \fancyfoot[R]{\includegraphics[width=90pt]{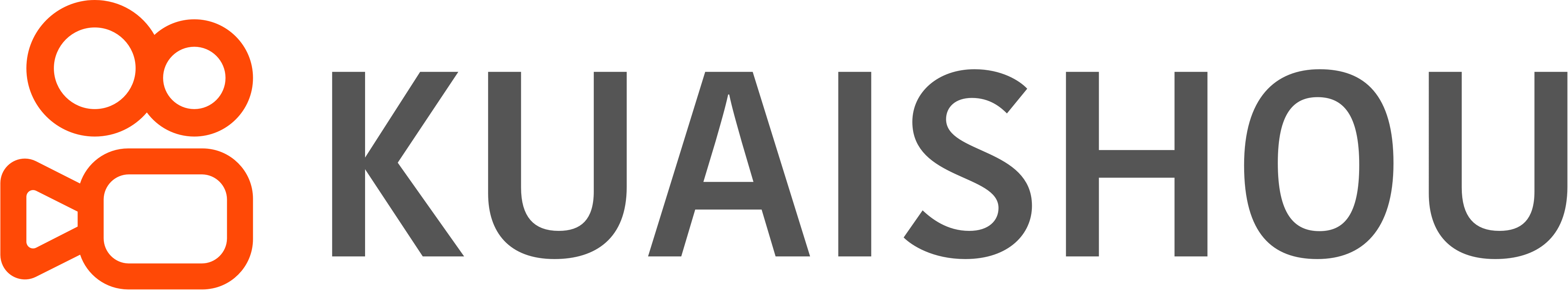}}
}

\begin{abstract}
Sustaining industrial recommendation research requires using the results of
one experiment to decide what to investigate next. We present AgentX-Model,
the next generation of AgentX's model research framework, which connects
proposal development and model experimentation within
sandboxes defined by business inputs and prediction tasks. AgentX-Model adopts
a dual-agent architecture comprising a Research Agent and a Model Agent.
The Research Agent develops independently reviewed proposals from papers and
experimental findings, while the Model Agent conducts multi-round investigations
and returns code, measurements, and unresolved questions. Using the returned
results, the Research Agent selects a starting implementation and formulates
the next research question, allowing subsequent experiments to build on earlier
findings. We organize this continuing research around four actions: Reproduce,
Follow-up, Composition, and Diagnose. The first three actions drive routine
research, while Diagnose acquires the evidence needed to choose a repair,
including for issues raised by business feedback and online evaluation,
such as prediction bias measured by PCOC. Across the
production evaluation, 560 of 636 completed model-changing
experiments recorded AUC above their business baselines. As research continued,
some experiments recorded AUC above every comparable ancestor in their
lineages. The five latest online A/B evaluations across different business
settings reported gains including 10--15\% in acquisition efficiency,
15--20\% in target-segment advertising spend, and 0.3--0.8\% in watch time;
the watch-time model used approximately 10\% fewer FLOPs and parameters.
A dependency-aware historical-replay benchmark further evaluates research
allocation, with initial results showing no consistent efficiency gain from
more complex scheduling when agents already analyze and select concrete candidates.
\end{abstract}

\begin{document}
\maketitle
\thispagestyle{agentxfirst}
\begingroup
\renewcommand{\thefootnote}{\fnsymbol{footnote}}
\footnotetext[1]{Equal contribution.}
\footnotetext[2]{Corresponding author.}
\endgroup
\clearpage
\begingroup
\setlength{\parskip}{0.4\baselineskip}
\tableofcontents
\endgroup
\clearpage

\section{Introduction}
\label{sec:introduction}

Industrial recommendation models develop through a succession of experiments.
An engineer introduces a method, examines the result, and decides whether to
refine the model, investigate an anomaly, or pursue another direction. Agents
now undertake increasingly complete parts of this work, from implementing
research requests to revising architectures with semantic
verification~\cite{autoreclab,nova}. Industrial systems also use asynchronous
experimentation and accumulated experience to support repeated model
changes~\cite{autorecsys,recevolve,astar}, and connect offline exploration with
deployment decisions~\cite{selfevolvingrec,autolr}. As agents take on this work, the
results of one experiment become material for deciding what to do next.

That transition involves choices that depend on the result. A useful model
may come from an intermediate round rather than the final revision. A
combination may need further work before it matches either source. A model
with higher offline AUC may expose a new problem when evaluated for business
use. Shared research records preserve results that later investigations can
build on~\cite{agora}. These cases raise a practical question: which
implementation, measurements, and open questions should guide the next experiment?

Our earlier work provides a foundation for this study. AgentX explored
paper-driven proposal generation, multi-round model research, and cross-paper
composition~\cite{agentx}. From Trajectories to Evidence studied how to associate
experimental conclusions with their code, measurements, and conditions, and
assess their use in a target experiment~\cite{trajectories}. AgentX-Model
builds on this foundation to carry experimental results into subsequent research.

AgentX-Model assigns model investigation and cross-experiment planning to
two agents. The Model Agent investigates code and training behavior through
multiple rounds; the Research Agent relates its findings to papers, business
feedback, and other experiments to develop independently reviewed proposals.
Their exchange lets a result become the starting point for another investigation.

\begin{figure}[!t]
\centering
\includegraphics[width=\textwidth]{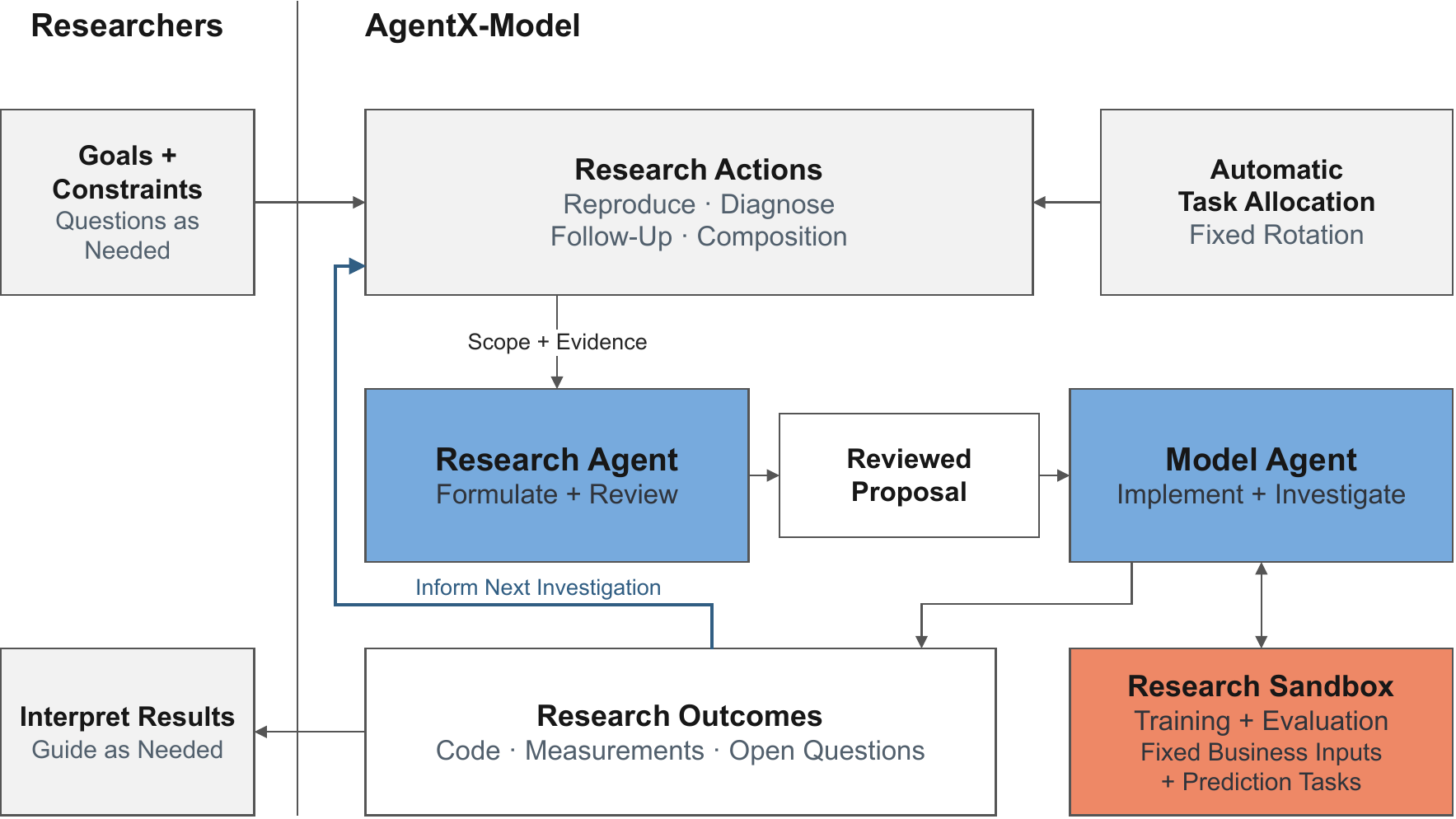}
\caption{Human--agent collaboration in AgentX-Model. The vertical line marks
the interaction boundary. Automatic allocation supplies a task type for the
Research Agent to consider. The blue return arrow feeds outcomes back into
research, so the right-hand loop can continue within the specified goals,
constraints, and resources without a new human request.}
\label{fig:research-cycle}
\end{figure}

In this technical report, we define autonomy as conducting and continuing research
within specified goals, constraints, and human-controlled deployment gates without a new human request
for every experiment. Within this setup, business inputs and prediction tasks
define the research sandbox. Researchers
set goals and constraints, introduce questions when needed, and interpret the
returned results. Within this scope, four actions express the research needs:
Reproduce introduces a method, Follow-up refines an implementation, and
Composition combines changes from different experiments. These three actions
drive routine research, while Diagnose investigates experimental observations
and business feedback to acquire the evidence needed to choose a repair.
Results return to researchers and feed the agents' next proposal
(Figure~\ref{fig:research-cycle}).

In production evaluation, 560 of 636 completed model-changing experiments
recorded AUC above their business baselines, and continued research produced new records
above all comparable ancestors in some lineages. The five latest online A/B
evaluations across different business settings reported gains including
10--15\% in acquisition efficiency,
15--20\% in target-segment advertising spend, and 0.3--0.8\% in watch time.
The watch-time model used about 10\% fewer FLOPs and parameters.

We also examine how to select the next experiment as research accumulates.
Like Dream-RSI~\cite{dreamrsi}, we use historical replay for this analysis,
preserving dependencies between experiments when comparing allocation policies.
Initial results suggest that, when agents already analyze candidate materials
and prior experimental results to select concrete investigations, more complex
adaptive or agent-based allocation does not consistently improve research
efficiency. Fixed task-type rotation combined with agent-based candidate
selection remains an effective baseline.

\begin{samepage}
This report makes three contributions:
\begin{itemize}
\item \textbf{A dual-agent framework for autonomous research.} We introduce
a Research Agent and a Model Agent with distinct responsibilities for
cross-experiment planning and within-experiment investigation, connecting
proposal development, multi-round experimentation, and research continuation.
Over approximately 25 days of observed production, the system completed
636 model-changing experiments.
\item \textbf{Four actions for long-horizon research.} We organize research
into Reproduce, Follow-up, Composition, and Diagnose, enabling the system to
introduce methods, refine implementations, combine findings, and revise
research questions in response to business feedback. In Scenario A, our
longest-running setting, 7 of 77 Follow-ups and 5 of 120 Compositions recorded
AUC above every comparable ancestor.
\item \textbf{A benchmark for continuing research decisions.} We construct
a dependency-aware historical-replay benchmark with 473 experiment nodes
across six environments to study how allocation policies and experience
reuse affect subsequent experiment selection.
\end{itemize}
\end{samepage}

\section{System Overview}
\label{sec:system}

AgentX-Model studies model changes within the inputs and prediction tasks of
a business setting. Within that scope, the two agents divide the work of
formulating a research question and investigating it. Their exchange of
proposals and results connects individual experiments into continuing research
paths (Figure~\ref{fig:research-cycle}).

\subsection{A Research Sandbox Defined by the Business Setting}

The business inputs and required predictions define the research sandbox.
Agents can change how the model uses those inputs, but do not change upstream
feature collection or redefine the prediction tasks. Training and evaluation
use the business data and execution environment. Every experiment is anchored
to a specified business baseline; its starting implementation may be that
baseline or a model retained from an earlier experiment. Both agents receive
the baseline's inputs, labels, outputs, and model structure, with code
references for the baseline and the chosen starting implementation.

Within this sandbox, agents can change feature representations and interactions,
routing, backbones, connections, and task towers. They can add, remove, replace,
merge, or tune components, and investigate losses or optimization where the
task permits. Each proposal specifies how the computation will change.
A replacement, for example, identifies both the new operation and the old
path to remove, so that it is tested as a replacement rather than implemented
as an additional branch.

The business baseline can evolve as models improve. Each result therefore
records the baseline version and evaluation conditions used in that experiment.
Findings from another business setting can suggest a useful direction, which
must then be adapted and tested with the target setting's inputs and prediction
tasks.

\subsection{Two Agents, Two Research Horizons}

The Model Agent follows the implementation closely, inspects training behavior,
and revises code in response to measurements. The Research Agent relates those findings
to other experiments, external papers, and business feedback. This separation
lets the Research Agent compare paths without carrying all training logs and
local tests in one conversation.

The agents coordinate through a \emph{proposal}: a document specifying a
research question and its experimental design. Each delivered proposal defines
one \emph{experiment}, which the Model Agent may investigate through multiple
\emph{rounds} of implementation and verification. Experiments connected by
inheritance or composition form a \emph{research path}. Proposal-production
attempts are counted separately from delivered experiments.

To begin an experiment, the Model Agent needs to know which code to modify,
what change to investigate, and how to judge the result. The proposal provides
these together with the research question. We summarize proposal $t$ as
\begin{equation}
P_t=(q_t,s_t,\delta_t,v_t),
\label{eq:proposal}
\end{equation}
where $q_t$ is the research question, $s_t$ the starting implementation,
$\delta_t$ the proposed modification or diagnostic intervention, and $v_t$
the evaluation design, including reference results and success criteria.
The Research Agent develops and reviews this proposal; the Model Agent
investigates it and returns the code and observations from its rounds.

Those returns form a shared research state for subsequent proposals:
\begin{equation}
S_t=(\mathcal I_t,\mathcal E_t,\mathcal Q_t),
\label{eq:research-state}
\end{equation}
where $\mathcal I_t$ contains implementations, $\mathcal E_t$ contains findings
tied to measurements and comparison conditions, and $\mathcal Q_t$ contains
unresolved questions. The implementations and findings retain their business
setting, baseline version, and evaluation context. When forming the next
proposal, the Research Agent can return to the code behind a result, examine
what has already been tried, and consider the questions that remain.

\section{Long-Horizon Research Cycle}
\label{sec:research-cycle}

\subsection{Conducting an Investigation}
\label{sec:refinement}
\label{sec:model-improvement}

Given an allocated research scope or a human question, the Research Agent's
Assembler investigates the sources and develops a proposal linking the proposed
change to the starting code and evaluation design.
An Auditor reviews it in a separate context, checking whether the change,
starting implementation, and evaluation together answer the research question.
The Assembler revises the proposal in response, or defers if it cannot support
the design. Source and version checks bind delivery to the approved proposal.
Appendix~\ref{app:proposal-refinement} details the review loop and its corrections.
Appendix~\ref{app:research-agent-evolution} follows the shift from prescribed
steps to agent-directed investigation.

The approved proposal gives the Model Agent a starting point $s_t$ and a scope
within which to investigate. It plans and codes a change, verifies the
implementation, and then trains and measures the model. The measurements can
motivate another revision or expose a question that needs further observation.
Each round returns its code and findings, retaining useful intermediate models
even when later revisions lose their gains. These returns update the shared
research state (Section~\ref{sec:evidence}) and make the next proposal possible:
\[
S_t \xrightarrow{\text{propose and review}} P_t
    \xrightarrow{\text{investigate and retain}} S_{t+1}.
\]
At each stage, the language model chooses what to inspect or revise; the
agent harness supplies tools and enforces execution limits.
Appendix~\ref{app:model-agent} details the implementation checks and feedback loop.

Some recurring difficulties concern the execution process rather than the
model being studied. The Model Agent can propose changes to tools or
instructions, validate them, and submit them for human approval before
deployment. It also autonomously judges whether an experiment offers reusable
experience and extracts candidate memories for human review; approved memories
guide later planning and coding. These changes leave the language model's
weights fixed, as in experience-based skill and memory
improvement~\cite{pilot,skillgen}. Related harness revision is studied in
HarnessDev and RobustSGPO~\cite{harnessdev,robustsgpo}.
Appendices~\ref{app:model-improvement} and~\ref{app:model-memory} describe the
two channels, and Appendix~\ref{app:model-evolution-results} evaluates an
agent-proposed, human-approved instruction revision on a fixed input batch.

\subsection{Continuing from Experimental Evidence}
\label{sec:evidence}

A continuation first needs an implementation to build on. The final version
is not always the useful one: a later change may have lost an earlier gain,
while a short summary may omit details needed to reconstruct the model that
produced it. Following the evidence-grounding approach of prior
work~\cite{trajectories}, we retain round-specific code with its measurements,
evaluation conditions, and diagnostic observations. Later revisions do not
overwrite the best measured implementation. Their failures and regressions
remain available too, recording what has been tried since that result.

Choosing the code does not yet determine how to judge the next experiment.
A repair may start from a diagnostic model while seeking to retain the ranking
gain of an earlier implementation. In that case, the diagnostic model supplies
the starting code, while the earlier ranking result supplies a performance
reference. We keep these choices separate in the proposal. The information
needed to make them is summarized in Table~\ref{tab:handoffs}.

\begin{table}[htbp]
\centering
\small
\caption{Information retained from an experiment and its use in subsequent research.}
\label{tab:handoffs}
\begin{tabularx}{\textwidth}{@{}YY@{}}
\toprule
Information carried forward & Decision in the next experiment \\
\midrule
Round-specific code and its baseline & Choose the implementation to resume from \\
Reference model, evaluation conditions, and measurements & Choose comparable reference results \\
Diagnostic observations and unresolved hypotheses & Design a test that distinguishes explanations \\
Complete effective changes from each source & Resolve overlap and design the joint modification \\
\bottomrule
\end{tabularx}
\end{table}

We distinguish three reference results: the \emph{business baseline} measures
benefit in the target setting; the \emph{direct parent} measures progress from
the chosen source experiment; and the \emph{strongest comparable ancestor}
tests whether the research path has reached a new best. For an AUC comparison,
let $\mathcal V_i$ be experiment $i$'s rounds with valid measurements under
the same evaluation conditions. Its best result is
\begin{equation}
b_i=\max_{r\in\mathcal V_i}\operatorname{AUC}_{i,r}.
\label{eq:best-round}
\end{equation}
Writing $\mathcal P_i$ for the comparable direct parents and $\mathcal A_i$
for all comparable ancestors, including those parents, gives
\begin{equation}
\begin{aligned}
\Delta_{\mathrm{base}}(i)   &= b_i-b_{\mathrm{base}(i)},\\
\Delta_{\mathrm{parent}}(i) &= b_i-\max_{j\in\mathcal P_i}b_j,\\
\Delta_{\mathrm{path}}(i)   &= b_i-\max_{j\in\mathcal A_i}b_j.
\end{aligned}
\label{eq:research-gains}
\end{equation}
Here $b_{\mathrm{base}(i)}$ is the recorded business-baseline AUC. Parent and
path gains are reported only when the required comparable results are
available; path gains require complete comparable ancestry. Repairing a
degraded descendant can make $\Delta_{\mathrm{parent}}$ positive while
$\Delta_{\mathrm{path}}$ remains negative.

The same attention to comparison applies when retaining findings in
$\mathcal E_{t+1}$. An improvement from $A$ to $A+B$ supports using the joint
model under the tested conditions; it does not measure $B$ independently.
New observations may support or challenge an earlier explanation and change
the questions in $\mathcal Q_{t+1}$. The next proposal can then choose an
implementation from $\mathcal I_{t+1}$ to investigate a remaining weakness or
its compatibility with another change. Papers and business feedback can also
introduce new directions. These questions determine which code to start from
and which references belong in the evaluation; diagnosis and calibration repair
need not begin with the highest-AUC ancestor (Section~\ref{sec:pcoc}).

\subsubsection{Four Research Actions}
\label{sec:actions}

The next research question depends on what is already known.
A new paper supplies a possible mechanism but little evidence in the target setting.
An anomaly supplies evidence of a problem but an incomplete explanation.
A concrete improvement suggestion supplies a direction for continued work.
Two successful paths supply candidate parts of a joint model. We represent
these needs as four research actions, with task-specific
starting points and interpretations of success.

\paragraph{Reproduce: introduce a mechanism into the target setting.}
Reproduce tests a paper-derived mechanism on a business baseline. The Research
Agent reads the source method, identifies the computation to test, and adapts
it to the available inputs and required predictions. When the experiment
transfers part of a method, the proposal identifies the retained mechanism and
the components outside its scope, defining the adaptation for the Model Agent
to implement and test.

\paragraph{Diagnose: acquire the evidence needed to choose a repair.}
When several explanations fit an observed failure, prescribing a fix can be
premature. A diagnostic task uses additional observations or targeted
interventions to test those explanations and guide a subsequent repair.
For example, a calibration task can vary the loss weight while tracking the
learned correction and prediction statistics to assess whether stronger
calibration pressure addresses the observed bias. The immediate objective is
to reduce a concrete uncertainty and establish what to investigate or repair next.

A Diagnose can follow a completed Reproduce, Follow-up, Composition, or
Diagnose experiment. Its implementation and findings can in turn support
a Follow-up, contribute a source to a Composition, or motivate another
Diagnose. These connections depend on the remaining question and available
results, allowing diagnosis to enter and continue a research path as needed.

\paragraph{Follow-up: continue a result with a specific purpose.}
A Follow-up starts from an implementation with measured results and investigates
an explicit problem or improvement opportunity. It carries the relevant code and evidence,
including the best version when later rounds have regressed. This gives the
Model Agent a defined research starting point while preserving its ability to
explore alternatives within the question. The target can be a higher AUC or a
business requirement such as calibration, with the appropriate performance
constraints retained.

\paragraph{Composition: test complementarity rather than accumulate modules.}
Two beneficial modifications can share an operation, replace the same module,
or depend on incompatible assumptions. The Research Agent therefore examines
their complete effective changes relative to the common baseline. Shared
changes are kept once, conflicts require a choice, and the proposal identifies
the distinct contribution remaining from each source. The experiment tests
whether those contributions work together. The goal is to exceed the best
comparable result from either source path, not just retain a gain over the
business baseline.

All four actions use the same proposal and execution interface, and each experiment
can contain several rounds of local investigation. They can also lead into
one another: a reproduced gain can prompt diagnosis, a diagnosis a Follow-up,
and a repaired result a later Composition.

A completed negative result can motivate a Follow-up if it exposes a specific
problem or a testable improvement. Infrastructure failures and executions
without valid measurements are excluded from automatic performance Follow-ups.
The path pauses when resources or a supported next question are unavailable;
its earlier valid implementations remain available for future research.

\subsection{Selecting the Next Investigation}
\label{sec:scheduling}

Several continuations and new methods may compete for the next experiment
budget, but their availability changes as research proceeds. A Composition
needs usable results from both source experiments; a Follow-up needs a source
implementation and a supported question. A task-type allocation still requires
deciding whether the evidence supports a concrete experiment.

\begin{figure}[htbp]
\centering
\includegraphics[width=\textwidth]{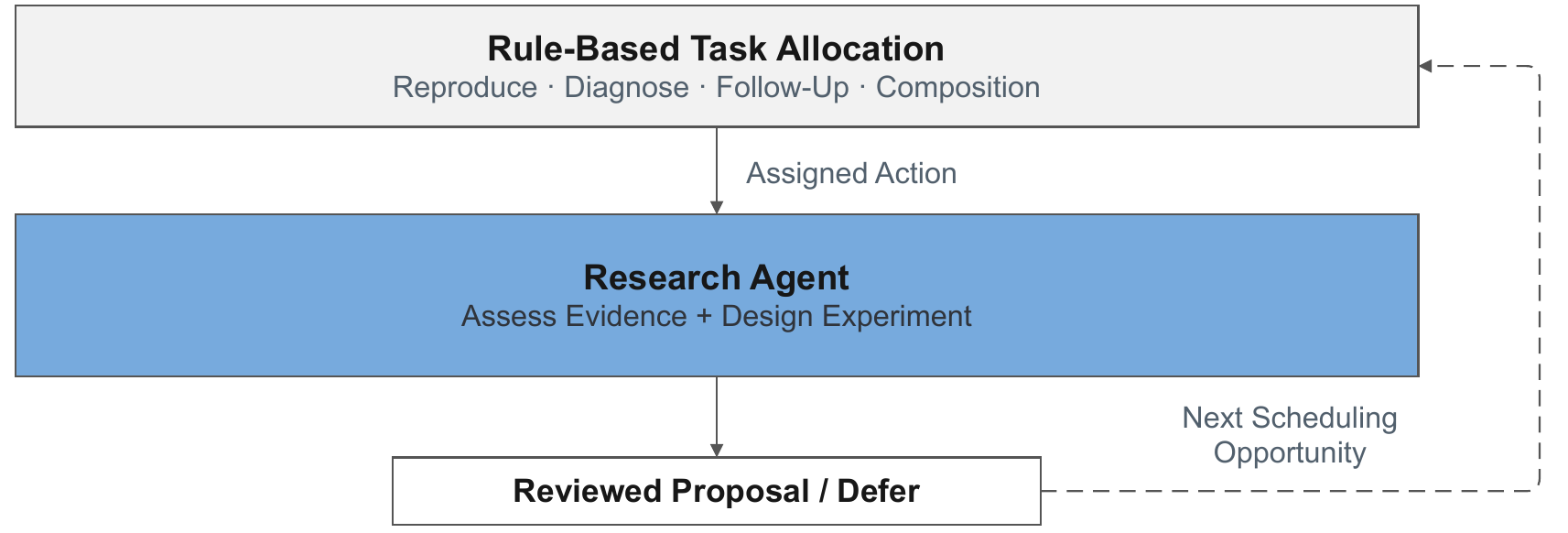}
\caption{Task allocation and research decisions. Human-designed rules allocate
a research-action type. The Research Agent uses available evidence to
construct and refine a concrete investigation or defer. The dashed return
marks the next scheduling opportunity.}
\label{fig:research-allocation}
\end{figure}

We use a human-designed scheduling policy as a baseline for sustained operation
and trajectory collection. We separate task-type allocation from concrete
proposal construction, a distinction also used in RecHarness~\cite{recharness}.
The policy allocates opportunities across Reproduce,
Diagnose, Follow-up, and Composition. Routine rotation and explicit research
demands enter the same allocation layer, which determines the action type to
consider (Figure~\ref{fig:research-allocation}).

Within this scope, the Research Agent decides what investigation the available
evidence supports. It examines the shared research state $S_t$, relevant
papers, and business feedback to select sources and a starting implementation,
formulate a research question, and design the experiment. When the evidence
supports a concrete experiment, the proposal enters the refinement loop in
Section~\ref{sec:refinement}; otherwise, the agent defers. Resource and
eligibility rules constrain the candidate pool and execution.
The resulting trajectories provide the candidate space and within-experiment
feedback for the scheduling benchmark in Section~\ref{sec:rq4}.

\section{Evaluation and Findings}
\label{sec:evaluation}

The evaluation follows the work from proposal production to model outcomes
and continued research. Production records show what was delivered and
measured; the resulting research paths show how implementations and questions
were taken forward. We then examine experience reuse and use historical replay
to compare ways of selecting subsequent experiments.

\subsection{Experimental Setting}
\label{sec:setting}

Scenarios A and D represent two distinct business settings; Scenarios B, C,
and E are baseline code versions within a third business setting.
The distinct code versions in B and E have the same measured baseline AUC
of 0.815403.
Production evaluation covers Scenarios A--D; historical replay also includes E.
We evaluate the models produced and refined within AgentX-Model against their
business baselines and online A/B controls.
Table~\ref{tab:evaluation-data} summarizes the production records and agent
evaluation datasets. The execution subset includes all 200 delivered proposals
with Research Agent execution records available in the initial frozen snapshot of the
sampled 10-day production records. The 189-experiment subset comes from the continuation
archive and requires at least two rounds with valid AUC per experiment.
We report online A/B results from the five latest Launch Reviews (LRs).

\begin{table}[!htb]
\centering
\small
\caption{Production records and agent evaluation datasets.}
\label{tab:evaluation-data}
\begin{tabularx}{\textwidth}{@{}>{\raggedright\arraybackslash}p{0.29\textwidth}>{\raggedright\arraybackslash}p{0.35\textwidth}>{\raggedright\arraybackslash}X@{}}
\toprule
Dataset & Size & Use \\
\midrule
\multicolumn{3}{@{}l}{\textbf{Production Records}} \\
Proposal production (10-day sample) & 886 attempts; 262 deliveries & Throughput \\
Offline results ($\sim$25 days) & 636 model-changing experiments & Baseline gains \\
Continuation archive & 1,218 experiments; 923 valid results & Research continuity \\
\midrule
\multicolumn{3}{@{}l}{\textbf{Agent Evaluation}} \\
Execution subset & 200 delivered proposals & Outputs and timing \\
Multi-round subset & 189 experiments & Best vs.\ final round \\
Historical replay & 473 nodes; 6 environments & Selection strategies \\
Instruction revision & 40 fixed inputs & Generation quality \\
\bottomrule
\end{tabularx}
\end{table}

Measurements are grouped by setting, baseline version, prediction head, and
evaluation procedure. Continuation comparisons use verified best-round AUC,
matching baseline entry-file content, prediction head, and reported baseline
AUC; strongest-ancestor comparisons also require complete comparable ancestry.
Each result table reports its eligible sample size.

With the random seed fixed, repeated executions yield pooled within-implementation
AUC standard deviations of 0.002208, 0.001010, and 0.001099 for Scenario A,
Scenarios B/C/E, and Scenario D, respectively. Under comparable evaluation
conditions, larger offline AUC gains relative to run-to-run variability help
prioritize candidates for online A/B tests that assess business impact.

\subsubsection{Implementation Details}
\label{sec:implementation}

During business production, the Research Agent used the Pi agent runtime,
with separate contexts for the Assembler and Auditor, while the Model Agent
used Claude Code. Both used privately deployed GLM-5.2 as the underlying
language model. The Model Agent submitted training and evaluation jobs to the
internal KML platform, retaining per-round code changes, measurements, and logs.
The Model Agent modifies model code, while evaluation uses fixed code provided
by the platform.
Subsequent historical-data processing, replay-based evaluation, and result
assessment used GLM-5.3.

\subsection{Sustained Operation in Production}
\label{sec:rq1}

We sampled the most recent 10 days of proposal production for this report,
recording 886 attempts and 262 proposals delivered to the Model Agent.
In the frozen subset of 200 delivered proposals, 198 had outputs
returned by the Model Agent and 177 yielded verifiable experimental measurements
(Table~\ref{tab:production}).

Producing these 200 proposals took a median of 33.6 minutes and a
90th percentile of 49.9 minutes, including investigation, review, and revision
but not downstream queueing or training.

\begin{table}[htbp]
\centering
\small
\caption{Proposal production in the sampled 10-day window. Execution
outcomes and production times are reported for a subset of 200 delivered proposals.}
\label{tab:production}
\begin{tabularx}{\textwidth}{@{}Yr@{}}
\toprule
Measure & Observation \\
\midrule
Proposal-production attempts & 886 \\
Proposals delivered to the Model Agent & 262 \\
Delivered proposals assessed for execution & 200 \\
Proposals with Model Agent outputs & 198/200 \\
Proposals with verifiable experimental measurements & 177/200 \\
Median proposal-production time & 33.6 min \\
90th-percentile proposal-production time & 49.9 min \\
\bottomrule
\end{tabularx}
\end{table}

\subsection{Model Gains and Research Continuity}
\label{sec:rq2}

The completed experiments provide both models to evaluate and starting points
for further work. We first compare their performance with business baselines
and report online A/B outcomes. We then follow how those models were revised
and combined, distinguishing performance inherited from earlier experiments
from new path-best records. A calibration case extends this analysis to a
business requirement that emerged after a ranking gain.

\subsubsection{Performance Gains Across Evaluation Settings}
\label{sec:business-gains}

\paragraph{Offline model performance.}
We summarize offline results from the latest stable production generation
over an approximately 25-day observation period.
Table~\ref{tab:offline-production-gains} reports each setting's baseline AUC,
best measured AUC, and the number of completed model-changing experiments
that exceeded their corresponding baseline. Diagnose experiments are excluded
from this model-gain summary; Section~\ref{sec:pcoc} illustrates how diagnosis
guides subsequent calibration repair.

\begin{table}[htbp]
\centering
\small
\caption{Offline model gains over approximately 25 days. Valid results are
completed experiments whose published AUC is verified against round-level
measurements. Best AUC is the highest measured AUC among these experiments;
$\Delta$AUC is its gain over the corresponding baseline. Results counted as
above business baseline exceed it by more than $10^{-6}$.}
\label{tab:offline-production-gains}
\begin{tabularx}{\textwidth}{@{}Yrrrr@{}}
\toprule
Evaluation setting & Baseline AUC & Best AUC & $\Delta$AUC & \shortstack[r]{AUC above business\\baseline / Valid (\%)} \\
\midrule
Scenario A & 0.773683 & 0.797352 & +0.023669 & 500/527 (94.9\%) \\
Scenario B & 0.815403 & 0.823145 & +0.007742 & 21/29 (72.4\%) \\
Scenario C & 0.816849 & 0.820246 & +0.003397 & 20/33 (60.6\%) \\
Scenario D & 0.809656 & 0.811432 & +0.001776 & 19/47 (40.4\%) \\
\bottomrule
\end{tabularx}
\end{table}

All four evaluation settings produced models above their business baselines.
The proportions in Tables~\ref{tab:offline-production-gains}
and~\ref{tab:offline-category-results} are not experiment success rates:
many Follow-up and Composition experiments inherit implementations that
already outperform the business baseline. Progress beyond inherited results
is assessed against parents and ancestors in the continuation analysis below.
The improvements involved different model changes. In Scenario B, the
best measured model expanded the expert pool from one to four and changed
the routing and load-balancing mechanism. In Scenario C, it restricted
information flow between token groups while adding a learned compensation
path. The best Scenario A result continued an implementation with accumulated
denoising, gating, and expert-routing changes, and corrected the scale of
its pairwise ranking loss. In Scenario D, a Follow-up addressed codebook collapse
by revising temperature normalization in the entropy regularizer.

\paragraph{Offline results by experiment category.}
Table~\ref{tab:offline-category-results} breaks down the same 636 completed
model-changing experiments by category; 560 exceeded their business baselines.
Knowledge Transfer is a proposal-source category: it supplies evidence from
another setting (Section~\ref{sec:knowledge-transfer-results}). The resulting
experiments follow the same proposal, execution, and review process.

\begin{table}[!ht]
\centering
\small
\caption{The same 636 completed model-changing experiments, grouped by category.
Definitions follow Table~\ref{tab:offline-production-gains}; percentages use
valid results within each category as the denominator.}
\label{tab:offline-category-results}
\begin{tabularx}{\textwidth}{@{}lYrrr@{}}
\toprule
Evaluation setting & Experiment category & \shortstack[r]{Valid\\results} & \shortstack[r]{AUC above business\\baseline} & \shortstack[r]{Share\\(\%)} \\
\midrule
Scenario A & Reproduce & 51 & 41 & 80.4\% \\
 & Composition & 308 & 302 & 98.1\% \\
 & Follow-up & 168 & 157 & 93.5\% \\
\midrule
Scenario B & Reproduce & 20 & 15 & 75.0\% \\
 & Follow-up & 9 & 6 & 66.7\% \\
\midrule
Scenario C & Reproduce & 24 & 15 & 62.5\% \\
 & Composition & 4 & 3 & 75.0\% \\
 & Follow-up & 5 & 2 & 40.0\% \\
\midrule
Scenario D & Reproduce & 7 & 0 & 0.0\% \\
 & Knowledge Transfer & 16 & 5 & 31.3\% \\
 & Composition & 2 & 1 & 50.0\% \\
 & Follow-up & 22 & 13 & 59.1\% \\
\bottomrule
\end{tabularx}
\end{table}

\paragraph{Online business impact.}
Table~\ref{tab:online-deployment-gains} summarizes the model changes and online
A/B results documented in these five LRs. Relative gains over the respective control
groups include acquisition efficiency, target-segment advertising spend, and
user engagement. The LR~3 model also reduced cost by pruning auxiliary training
objectives unused in online prediction, while retaining the required outputs.

\begin{table}[!ht]
\centering
\small
\caption{Online A/B results from the five latest Launch Reviews.
Relative gains use coarse ranges.}
\label{tab:online-deployment-gains}
\begin{tabularx}{\textwidth}{@{}l>{\hsize=0.8\hsize}Y>{\hsize=1.2\hsize}Y@{}}
\toprule
LR & Model change & Relative improvement \\
\midrule
1 & Adaptive attention temperature & Acquisition efficiency: 10--15\% \\
\addlinespace
2 & Attention-module combination with normalization & Target-segment advertising spend: 15--20\% \\
\addlinespace
3 & Multi-task objective pruning & Watch time across two clients: 0.3--0.8\%; FLOPs and parameter count: approximately 10\% lower \\
\addlinespace
4 & Hierarchical aggregation with calibration & Overall daily active users: 0.5--1\%; target-page follow actions: 3--4\% \\
\addlinespace
5 & Cross-setting mechanism transfer & Target-segment advertising spend: 5--10\% \\
\bottomrule
\end{tabularx}
\end{table}

\subsubsection{Evolution Through Follow-up and Composition}
\label{sec:continuation-results}

The baseline comparisons describe the models obtained. To understand how
research continued, we now follow the proposals and results over time, examine
one branching lineage, and compare continuations with their parents and
ancestors.

\begin{figure}[!t]
\centering
\includegraphics[width=\textwidth]{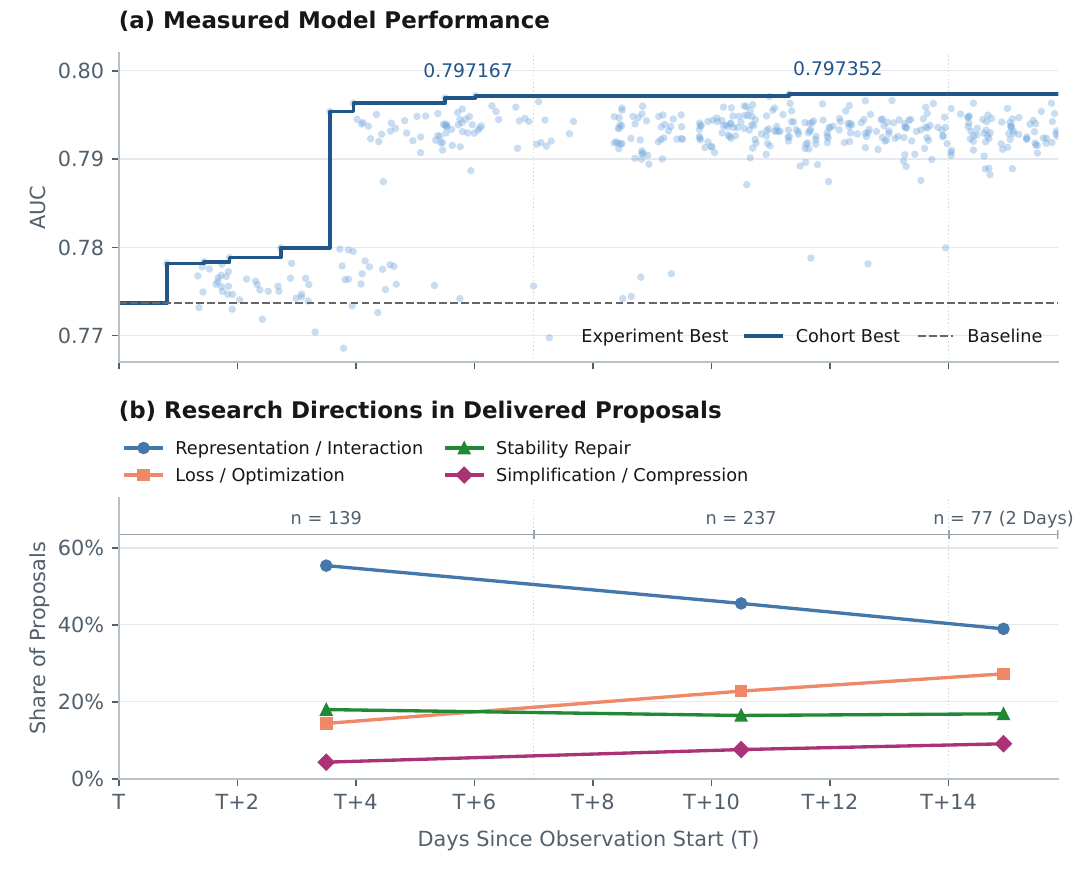}
\caption{Model and research evolution from a fixed baseline in Scenario A. (a) Points show experiment-best AUC by result-report time; the step
line tracks the cohort best from the business baseline. Experiments may
inherit implementations from before the delivery window.
(b) Shares use all proposals in each interval as the denominator, with one
primary direction per proposal; four of seven directions are shown.
$T$ marks the observation start. The final interval spans only two days,
$T+14$ through $T+15$.}
\label{fig:scenario-a-research-evolution}
\end{figure}

\paragraph{The best AUC record levels off as research questions shift.}
\looseness=-1
From the production records underlying Table~\ref{tab:offline-production-gains},
we track 453 Scenario A proposals delivered under one fixed business baseline
over the final 16 calendar days: 317 Compositions and 136 Follow-ups.
Of these 453 delivered proposals, 432 yielded valid completed results and are
included in the offline summary above.
Figure~\ref{fig:scenario-a-research-evolution} places their
measured performance alongside the changing distribution of proposal topics.
An LLM assigned each proposal a primary direction based on its research
question and proposed modification. Labels describe the new change rather
than inherited modules; secondary directions were retained in the annotations.
The cohort's cumulative best AUC reached 0.797167 on day $T+6$, then
increased by 0.000185 to 0.797352 on day $T+11$ and remained unchanged
through day $T+15$, where $T$ marks the observation start. Meanwhile, representation and interaction questions
declined from 55.4\% to 39.0\% of delivered proposals, while loss and
optimization rose from 14.4\% to 27.3\% and simplification from 4.3\% to
9.1\%. The best AUC record changed little in the later part of the observation
window, while a growing share of proposals examined training objectives and
model simplification.

\begin{figure}[!htb]
\centering
\includegraphics[width=0.9\textwidth]{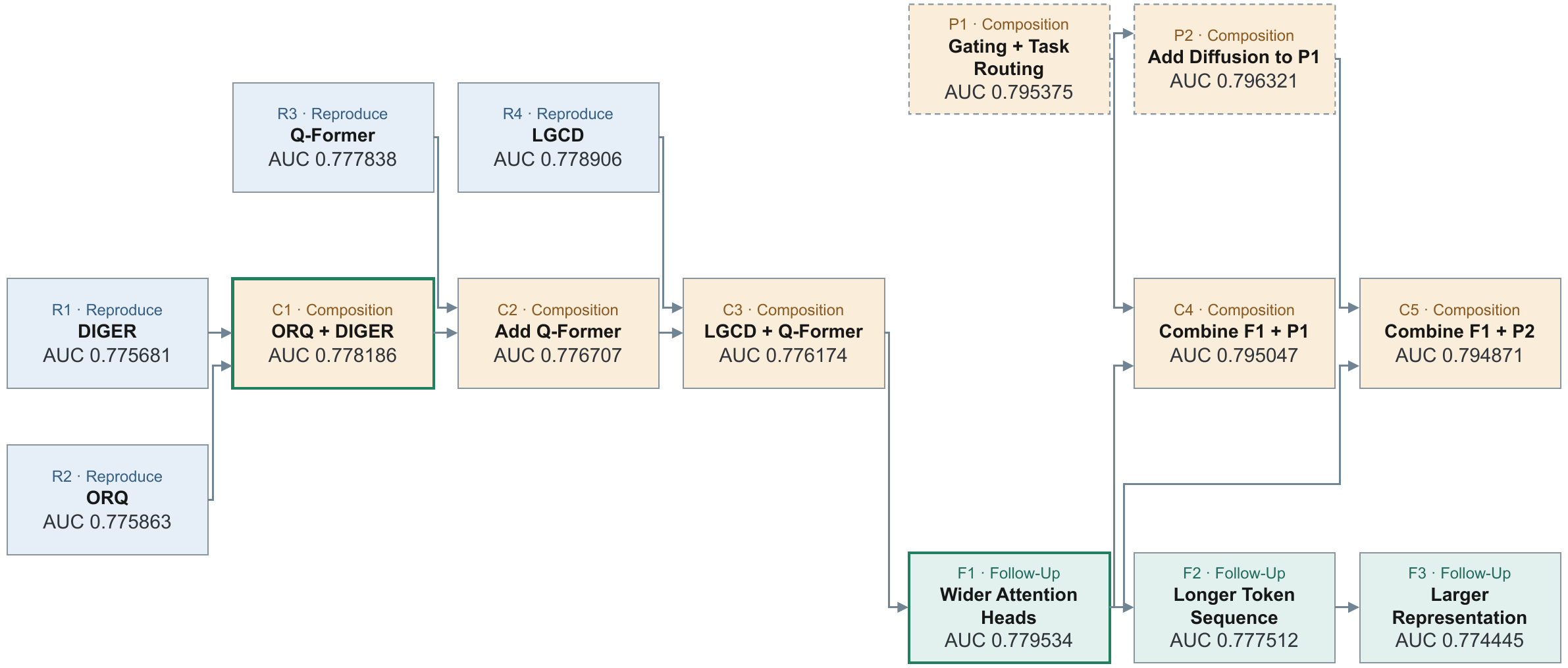}
\caption{An example research lineage spanning approximately 11 days in Scenario A.
Each node is an experiment labeled with its best measured AUC; arrows
connect parents to children. Green borders mark new path-best AUC records.
Dashed boxes summarize the other path's earlier ancestry.}
\label{fig:scenario-a-lineage}
\end{figure}

\paragraph{One result becomes the starting point for several branches.}
Figure~\ref{fig:scenario-a-lineage} shows a 14-experiment excerpt from a
Scenario A research lineage under a fixed business baseline (AUC 0.773683).
It spans approximately 11 days from the first proposal delivery to the final
returned result. Four paper-derived Reproduce
experiments feed successive Compositions. Follow-up F1 then records the highest
AUC along its ancestry and becomes a source for three continuations: another Follow-up
and two Compositions with implementations from a different research path.
The Follow-up branch records lower AUC in F2 and F3. Compositions C4 and C5
combine F1 with P1 and P2 but remain below those stronger parents by
0.000328 and 0.001450 AUC.

\paragraph{Comparing continuations with parents and ancestors.}
The lineage example shows why a continuation has more than one relevant
reference: its immediate source may differ from the strongest earlier model.
We examine this distinction across experiments using the separate frozen
continuation archive.
In the Scenario A records, 46 of 91 comparable Follow-ups have a higher
best-round AUC than their direct parent. Among the 77 Follow-ups with complete
comparable ancestry, 7 exceed every ancestor (Table~\ref{tab:continuation}).
Composition also yields few new path-best records. The parent and ancestor
comparisons distinguish a local recovery from a new record along the path.

\begin{table}[htbp]
\centering
\small
\caption{Scenario A continuation results. Each cell gives experiments with
a higher best-round AUC over the number with the required comparable evidence.
An increase must exceed $10^{-6}$ to avoid rounding ties. Complete ancestry
requires more evidence, so the column denominators differ.}
\label{tab:continuation}
\begin{tabularx}{\textwidth}{@{}Yrr@{}}
\toprule
Research action & \shortstack[r]{AUC above strongest\\direct parent} & AUC above all ancestors \\
\midrule
Follow-up & 46/91 (50.5\%) & 7/77 (9.1\%) \\
Composition & 6/139 (4.3\%) & 5/120 (4.2\%) \\
\bottomrule
\end{tabularx}
\end{table}

\paragraph{Preserve the best measured implementation, not just the last.}
The comparisons above use each experiment's best measured round. A later
experiment needs access to the code from that round, which may differ from
the final implementation. Among the 189 experiments with multiple
valid round-level measurements, 111 (58.7\%) record higher AUC after the first valid
round, but 91 (48.1\%) have a last valid round whose AUC falls below an earlier
best. These groups can overlap: an experiment can exceed its first-round AUC
and still finish below its intermediate best. Here, ``last''
means the last round with a valid AUC measurement, not a subsequent failed
attempt.
Keeping the best measured code with its evaluation context preserves that
starting point. Lower-scoring rounds and failures record subsequent attempts
and inform the next research question.

\paragraph{A combination that required a further model revision.}
In a separate Scenario A Composition, the first source had adapted language-guided conditional
diffusion (LGCD)~\cite{lgcd} to the model's feature-fusion path and reached 0.778906.
The second was a previous Composition whose shared variant of the
Attention-Free Token Mixer (AFTM) from FuXi-$\beta$~\cite{fuxibeta}
reached 0.775554. Its best round disabled the additional
traffic-specific branch. The Research Agent selected that shared mixing
operation for the new Composition rather than carrying over the entire
second model.

\begin{figure}[htbp]
\centering
\includegraphics[width=\textwidth]{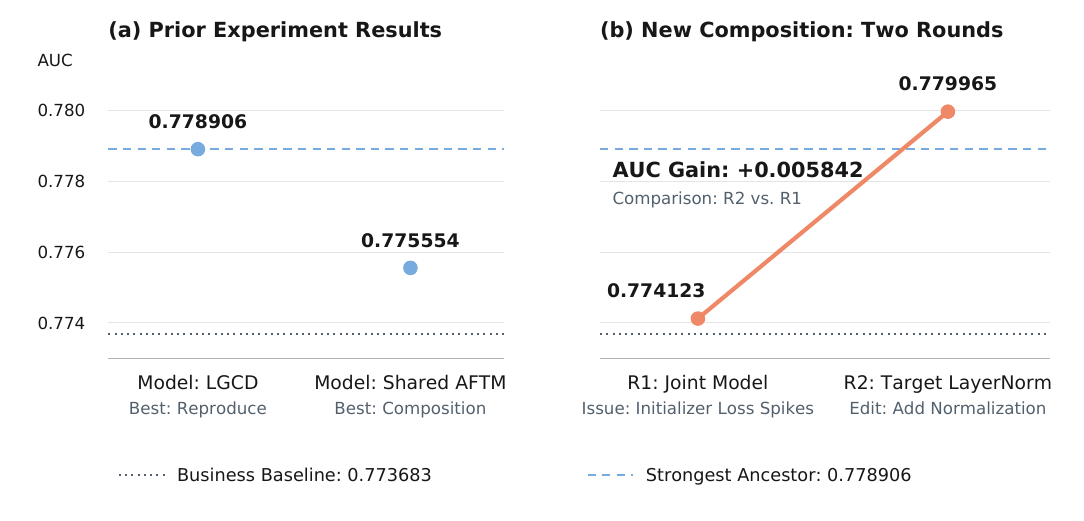}
\caption{A Scenario A Composition followed by within-experiment refinement. The left panel
shows the best rounds of the two direct source experiments, not successive
rounds. The right panel shows the two rounds of the new joint model. The LGCD
source is the new experiment's strongest comparable ancestor. Dashed reference
lines mark the business baseline and strongest-ancestor AUC; only the two
Composition rounds are connected by a line.}
\label{fig:composition-case}
\end{figure}

The proposal replaced cross-attention in the conditional diffusion denoiser
with SiLU-gated attention-free mixing. The Model Agent built a joint model
with conditional generation and mixture-of-experts fusion.
Figure~\ref{fig:composition-case} separates the two source results from the
rounds of this new experiment.

The first round reached 0.774123, below both direct sources. Its report
identified spikes in the loss used to train the diffusion initializer. The
Model Agent then normalized each diffusion-target token with a non-affine
LayerNorm. Comparing the two rounds' code changes confirms that this local
normalization was the only implementation change between them. The second
round reached an AUC of 0.779965, 0.005842 above the first round and 0.001059
above the strongest ancestor. Its report described smaller loss spikes after
the revision. The training problem exposed
by the initial combination prompted this local revision.

\subsubsection{From Ranking Gains to Calibration Repair}
\label{sec:pcoc}

Further research can also be prompted by a business requirement rather than
another structural change. In Scenario A, we report AUC for ranking
performance and PCOC for aggregate calibration.
An adaptation of SMES~\cite{smes}, which routes each prediction task to
selected shared expert networks, improved offline AUC from
0.773683 to 0.778940. Business feedback then identified prediction bias,
giving the next investigation a specific purpose: improve
calibration while preserving the ranking gain. PCOC is the ratio of predicted
to observed totals:
\begin{equation}
\operatorname{PCOC}=\frac{\sum_n\hat y_n}{\sum_n y_n},
\label{eq:pcoc}
\end{equation}
Here $\hat y_n$ and $y_n$ are the prediction and observed label for sample $n$.
PCOC above 1 indicates overprediction and below 1 underprediction.

Business review identified online overprediction for the original model and
an offline--online calibration mismatch for an earlier repair.
These observations motivated the offline Diagnose and immediate
Follow-up described below. The Research Agent began with a hypothesis about
the existing correction: would a larger calibration-loss weight move the
learned correction further and improve PCOC? It organized a Diagnose to
examine that possibility.

The Diagnose increased the calibration-loss weight from 1 to 10 and used a
direct SMES replacement, whereas the earlier repair used a layer-normalized
gated residual connection. In this configuration, the learned correction
parameter, the log-bias, remained near $-0.10$, close to the historical value
of $-0.09$, and test PCOC was 0.953836. Training observations led the Model
Agent to infer that the parameter had settled under the current objective.
It proposed examining the calibration statistics next, rather than increasing
the loss weight again.

The Follow-up pursued a correction estimated from the calibration statistics.
During proposal review, the Auditor caught a draft targeting a different
prediction task and required the original task and Diagnose
implementation to be inherited explicitly (Appendix~\ref{app:auditor-cases}).
The reviewed proposal used the Diagnose code and estimated a multiplicative
correction from training data. For the training samples $\mathcal T$
accumulated after warm-up, the correction factor was
\begin{equation}
c=\frac{\sum_{n\in\mathcal T}y_n}{\sum_{n\in\mathcal T}\hat y_n},
\label{eq:pcoc-correction}
\end{equation}
using predictions before the new correction and their corresponding
labels. The factor was applied to additional evaluation outputs, without
fitting to test labels or changing the existing inference outputs.
The first round collapsed during training, with
an AUC of 0.500000. The Model Agent reran the same code and obtained
two healthy executions (Table~\ref{tab:pcoc}).

\begin{table}[htbp]
\centering
\small
\setlength{\tabcolsep}{5pt}
\caption{The Scenario A calibration path. Before/after values compare outputs of
the same run, retaining the inherited calibration and then adding the new
factor. Error reduction is relative to $|\mathrm{PCOC}-1|$ before correction.
All Follow-up rounds use the same code patch; R1 collapsed during training.}
\label{tab:pcoc}
\begin{tabularx}{\textwidth}{@{}Yrrrr@{}}
\toprule
Implementation & AUC & PCOC before & PCOC after & Error reduction \\
\midrule
Business baseline & 0.773683 & -- & -- & -- \\
SMES best & 0.778940 & -- & -- & -- \\
Diagnose & 0.775103 & 0.953836 & -- & -- \\
Follow-up R1 & 0.500000 & -- & -- & -- \\
Follow-up R2 & 0.775632 & 0.951715 & 0.962735 & 22.8\% \\
Follow-up R3 & 0.775318 & 0.948190 & 0.959965 & 22.7\% \\
\bottomrule
\end{tabularx}
\end{table}

Correction factors estimated from training data reduced test-set calibration
error by 22.8\% and 22.7\% in the two healthy runs.
Their ranking AUC observations remained above the business baseline.

The path began with a ranking gain and continued with a calibration question
raised by business feedback. It motivates \emph{staged optimization with joint
acceptance}: first find a useful ranking model, then require the final model
to meet business constraints while retaining its ranking benefit.

\subsection{Knowledge Transfer Across Settings}
\label{sec:rq3}
\label{sec:knowledge-transfer-results}

The preceding cases continue research within a business setting. We also
explored whether findings from one setting could provide starting points in
another. The source settings offered more room for structural improvement at
lower iteration cost, whereas Scenario D had a more heavily optimized starting
model and required more resources and time for each trial. These differences
motivated accumulating findings in the lower-cost settings and using them to
formulate proposals for the more costly target setting.

We call this cross-setting proposal route \emph{Knowledge Transfer}.
The Research Agent reads source-experiment code and results and compresses
them into guidance describing the tested changes, observed outcomes, and
conditions relevant to reuse. It then examines the Scenario D baseline and
develops a proposal specifying how to adapt and test the selected idea with
the target inputs and prediction tasks. The proposal goes through independent
review before the Model Agent implements and evaluates it in Scenario D.
Paper-derived proposals instead take their candidate mechanisms directly
from a paper's method. The comparison thus concerns where the evidence for
a new research question comes from: a published method or findings from
experiments in another setting.

In this exploratory comparison, we examine the two proposal sources in the
frozen continuation archive.
The two historical cohorts were collected in different periods without
matched experiment budgets.
Restricting the comparison to matching baseline entry-file content, the same
prediction head, and baseline AUC of 0.809656 leaves seven paper-derived
experiments and 15 Knowledge Transfer experiments with valid published AUC.
The broader production summary in Table~\ref{tab:offline-production-gains}
includes one additional measured Transfer whose starting entry-file content
could not be verified.

\begin{table}[htbp]
\centering
\small
\setlength{\tabcolsep}{9pt}
\caption{Historical Scenario D results by proposal source. $\Delta$AUC is the
published AUC minus the common baseline AUC of 0.809656.}
\label{tab:knowledge-transfer-results}
\begin{tabular}{@{}lrr@{}}
\toprule
Proposal source & Valid results & $\Delta$AUC $>0$ \\
\midrule
Paper-derived & 7 & 0 (0\%) \\
Knowledge Transfer & 15 & \textbf{5 (33.3\%)} \\
\bottomrule
\end{tabular}
\end{table}

In this cohort, Knowledge Transfer produced more above-baseline AUC
observations: five of 15 experiments, compared with none of seven paper-derived
Reproduce experiments (Table~\ref{tab:knowledge-transfer-results}). The largest
increment was 0.000789.
These above-baseline implementations provide more candidate starting points
for subsequent research.

\subsection{Selecting the Next Experiment}
\label{sec:rq4}

Forming a useful research proposal leaves a further choice when several
directions are available: which experiment should run next? We study this
choice through historical replay, where policies select among recorded
experiments and receive their results. The benchmark compares fixed,
feedback-driven, and agent-led allocation by the results found, selections
required, and decision cost. It also examines stalled searches and, in a
separate comparison, whether summaries help an agent that can already inspect
source code and results.

\paragraph{Dependency-aware graph replay.}
Nodes represent experiments and edges represent their recorded prerequisite
dependencies. Each replay run starts from the business baseline with no
experiment nodes preselected. Follow-up and Composition candidates become
available after their prerequisite experiments have been selected. A Composition requires
multiple source experiments, so its availability depends on progress along
more than one research branch. Selecting an experiment therefore both reveals
its results and can unlock subsequent candidates. The replay captures this
changing opportunity set when comparing selection policies.
Each selection counts once against the experiment budget and returns the
selected experiment's recorded rounds as feedback. Outcomes are available
only for combinations that were actually executed.

We construct separate replay environments for each baseline version.
Scenario C has a baseline AUC of 0.816849 and Scenario E has 0.815403.
The replay evaluation contains six test environments with 473 experiment
nodes: one graph each for Scenarios A, B, C, and E with 348, 26, 21, and
69 nodes, respectively, and two Scenario D graphs with three and six nodes.
Scenario C includes two Compositions, whereas Scenario E has none.

Before selection, agents see opaque candidate identifiers, action types, and
links to revealed parents and sources, and can inspect the baseline code and
an eligible candidate's first implementation, its pre-change code, and their
diff. This view contains neither original proposal text nor post-execution
reports; the candidate's measurements and later revisions remain hidden.
Selecting it reveals recorded round-level measurements, execution and review
status, failure reasons, and best-round comparisons, and unlocks available code
from its recorded rounds for subsequent decisions.

\paragraph{Experiment selection policies.}
The three main policies share the LLM configuration described in
Section~\ref{sec:implementation} and use the same tools, eligibility conditions,
and experiment budget. \emph{Fixed routing} cycles through Composition,
Follow-up, Follow-up, and Reproduce, skipping unavailable types; the agent
selects a candidate within the chosen type. \emph{Bandit routing} adapts the
type allocation from revealed outcomes and uses the same candidate selector.
\emph{Joint selection} lets the agent choose the type and candidate together.
The comparison therefore changes how opportunities are allocated while
retaining agent-based candidate selection in all three policies. Fixed routing
serves as the baseline for comparing allocation policies in historical replay.
The recorded opportunities reflect prior human-designed and agent-assisted
selection, often extending promising paths, and cover only branches that were
executed. The comparison is therefore conditioned on this historical sampling.

The routing comparison on Scenarios A, B, C, and E includes Reproduce, Follow-up, and Composition.
Diagnose is excluded because the value of information for a later repair is
not captured by immediate AUC. Bandit routing uses Thompson Sampling with a
Beta$(1,1)$ prior, rewarding a strict new highest valid AUC among revealed
results, including the baseline. We also include Uniform Random and
parent-result Greedy as low-cost references. Greedy ranks legal candidates
by revealed direct-parent best AUC, substitutes the baseline for roots or
missing parent measurements, and breaks ties with a seeded random generator.

Each replay run allows at most 20 experiment selections, with checkpoints at
5, 10, and 20. It stops earlier if no candidates remain. We run all five
policies three times on each of the four graphs for Scenarios A, B, C, and E, for 60 runs. The two
small Scenario D graphs are evaluated only with Uniform Random and parent-result
Greedy, each repeated three times, for another 12 runs. Agent sessions and
policy state are reset between repeats. All 72 runs completed.

A run reaches the target when it first finds a valid AUC strictly greater than
the comparable baseline plus 0.001. For runs that reach the target, we report
the number of selections required. Attainment and completion rates use all
planned runs as their denominator.
We report best AUC after 20 selections, decision time, and token use.
The same number of selected experiments can represent different historical
training costs.

\paragraph{Results found across environments.}
Final AUC rankings varied across replay environments, and no allocation policy
consistently led across them (Table~\ref{tab:replay-final}). Fixed rotation
with agent-based candidate selection remained competitive.

Attainment of the predefined baseline-plus-0.001 target gives a first view
of what the policies found.
All 60 runs on Scenarios A, B, C, and E reached it, including
46 on the first selection. This threshold therefore provides limited
separation among policies on these graphs. Neither Scenario D graph contains a
result that meets it: the six-node Transfer graph reaches 0.810445 from a
baseline of 0.809656, a gain of 0.000789, while the three-node Reproduce graph
has no result above that baseline. The 12 Scenario D runs exhaust their small
candidate pools without attaining the target.

The highest AUC records after 20 selections distinguish the policies further.
In Scenario A, Uniform Random's mean exceeds Fixed, Greedy, and Joint.
Bandit has the highest mean on the two larger
graphs, but its Scenario A mean is strongly influenced by one run reaching 0.788360;
the sample standard deviation across its three Scenario A runs is 0.005092.
On the Scenario B and C graphs, final scores are identical
or nearly identical, which leaves a different question: how many selections
were needed to reach them?

\begin{table}[htbp]
\centering
\small
\setlength{\tabcolsep}{4pt}
\caption{Mean best AUC after 20 selections on the four graphs for Scenarios A, B, C, and E.
Each policy completes three runs per graph; baselines and evaluation contexts
are row-specific. Fixed and Bandit allocate the type before agent candidate
selection; Joint selects both.}
\label{tab:replay-final}
\begin{tabularx}{\textwidth}{@{}Yrrrrrr@{}}
\toprule
Evaluation setting & Baseline & Random & Greedy & Fixed & Bandit & Joint \\
\midrule
Scenario A & 0.773683 & 0.782285 & 0.779145 & 0.780657 & 0.782500 & 0.780012 \\
Scenario B & 0.815403 & 0.823145 & 0.823141 & 0.823141 & 0.823141 & 0.823141 \\
Scenario C & 0.816849 & 0.819780 & 0.819780 & 0.819780 & 0.819780 & 0.819780 \\
Scenario E & 0.815403 & 0.821067 & 0.820931 & 0.821011 & 0.821614 & 0.821011 \\
\bottomrule
\end{tabularx}
\end{table}

\paragraph{Selections required to reach the same result.}
Final AUC can conceal differences in the number of experiments required to
reach it. The Scenario C graph has 21 nodes, so the 20-selection
budget covers nearly the entire graph. All five policies reached the same
highest recorded AUC of 0.819780 in all three repeats.
This result is a Composition of two Reproduce experiments whose best AUCs
are 0.818920 and 0.819011, both above the 0.816849 business baseline.
The two sources are parallel research paths: both must be selected before
their combination becomes available. The minimum is three selections; the 15
runs reached the combination after 7--20 selections.

Table~\ref{tab:replay-attainment} provides a post-hoc comparison of the
selections required to reach this shared endpoint, including all prerequisites.
Across three repeats on Scenario C, Joint reached the shared endpoint after
10.67 selections on average (range 7--16), compared with 14.67 for Fixed
(12--16). Both policies reached the same model through different sequences
of prerequisite and other experiments.

\begin{table}[htbp]
\centering
\small
\caption{Selections to first reach the highest recorded AUC (0.819780) on
the Scenario C graph. Each policy has three repeats; all 15 runs
reach this result within the 20-experiment budget.}
\label{tab:replay-attainment}
\begin{tabular*}{\textwidth}{@{\extracolsep{\fill}}lrrr@{}}
\toprule
Policy & Mean & Median & Range \\
\midrule
Uniform Random & 18.00 & 18 & 16--20 \\
Parent-result Greedy & 13.33 & 15 & 8--17 \\
Fixed routing + Agent & 14.67 & 16 & 12--16 \\
Bandit routing + Agent & 16.67 & 17 & 15--18 \\
Joint selection & 10.67 & 9 & 7--16 \\
\bottomrule
\end{tabular*}
\end{table}

\begin{samepage}
\paragraph{When subsequent experiments remain unavailable.}
Candidate availability exposed a different obstacle in Scenario A. The graph
contains 83 Compositions, yet none was selected in the 15 runs. In all nine
agent runs, no Composition became eligible because its required source
experiments had not all been selected. Preferring Composition therefore did
not suffice to make a combination executable. Whereas the Scenario C policies
eventually reached the combination, these Scenario A runs never unlocked one.
This contrast motivates evaluating an experiment not only by its recorded
result, but also by the subsequent research opportunities it makes available.

\end{samepage}

To examine whether other stalled searches left opportunities unused, we ask
at checkpoints 5 and 10 whether an unselected better
result remained reachable in the frozen graph. Its minimum additional cost is one selection
for that result plus its unfinished dependencies, counting shared dependencies
once. This distinguishes three situations: an improvement reachable within
budget, an improvement requiring more budget, or no better reachable recorded
result. The analysis uses hidden outcomes after the run; the selection policy
does not see them. This dependency-based lower bound does not include any
additional selections imposed by fixed task-type rotation.

Across 120 checkpoints from the 60 runs on Scenarios A, B, C, and E, 110 still had a better recorded result
reachable within the remaining budget. Of those 110 checkpoints, 19 were
followed by no higher AUC record. The remaining 10 of the 120 checkpoints had no better
reachable record; none was limited solely by the cost of unfinished
dependencies. A flat search curve can therefore reflect either
an unused opportunity or the limit of the recorded pool.

\paragraph{Decision cost.}
The 36 agent runs took approximately 430 minutes in total: about 124 for
Fixed, 157 for Bandit, and 149 for Joint. They used 97.5 million metered tokens, including
86.7 million cache-read tokens. These costs measure selection and evidence
inspection in replay, separately from the historical experiments' training
costs. Uniform Random and parent-result Greedy required no language-model calls.
Their competitive results provide low-cost references for replay.

\paragraph{Selecting with summarized experience.}

Allocation determines which opportunities the agent considers; the evidence
it receives then informs the choice. In an exploratory comparison, we hold
the Joint selector fixed and
compare access to raw source-experiment code and AUC measurements with access
to the same material plus compressed summaries.

In each of Scenarios A and E, six development experiments supply three
source-reviewed summaries. An initial batch emphasizes actionable modifications
and reuse hypotheses; a second batch uses the same sources but specifies the
intervention location, comparison, measurement, and conditions for reuse.
Summary generators could access only the development-source
code and measurements, not target candidates or their results. Source reviewers
used these materials to correct factual errors, but already knew results from
earlier benchmark runs on the same target graphs; the review was not blind.
Each summary batch was frozen before its paired selection runs.

We retain the Joint selector's model, tools, and 20-selection budget from the
routing study. Each batch has three paired repeats per setting, giving two
12-run batches reported separately from the 72-run routing comparison.
Two of the three initial Scenario A pairs completed all 20 selections.
In the third pair, the summary condition timed out after 18 selections.
Table~\ref{tab:knowledge-summary-selection} includes the two complete pairs
for that row; every other row includes three complete pairs.

Alongside best AUC and selections to the same result, we examine a post-hoc
measure of candidate choice: how often the selected experiment has a positive
recorded gain over its baseline. This criterion is $\Delta$AUC $>0$, not the
routing study's attainment threshold of $0.001$ or a gain over the parent.
Counts pool replay selections across repeats, which can revisit the same
historical experiment. The excluded partial pair gives 16 positive selections
in each condition over their common first 18 selections.

\begin{table}[htbp]
\centering
\small
\setlength{\tabcolsep}{5pt}
\caption{Summary comparisons in historical replay. Above-baseline selections count
experiments whose best valid AUC exceeds their baseline, summed across
repeats. Paired $\Delta$Best@20 is with-summary minus raw-evidence best AUC.
The initial Scenario A row includes two complete pairs; each other row includes three pairs.}
\label{tab:knowledge-summary-selection}
\begin{tabular}{@{}llrrl@{}}
\toprule
 & & \multicolumn{2}{c}{Above-baseline selections} & \\
\cmidrule(lr){3-4}
Summaries & Evaluation setting & Raw & + Summary & Paired $\Delta$Best@20 \\
\midrule
Initial & Scenario A & 37/40 & 37/40 & $+0.000998,\;-0.000950$ \\
Initial & Scenario E & 51/60 & 55/60 & $0,\;0,\;-0.001809$ \\
Comparative & Scenario A & 54/60 & 52/60 & $-0.001052,\;+0.000329,\;+0.017284$ \\
Comparative & Scenario E & 49/60 & 51/60 & $0,\;0,\;-0.001602$ \\
\bottomrule
\end{tabular}
\end{table}

The summaries reduce selections with non-positive recorded gains in both
Scenario E batches. In Scenario A, positive selections are unchanged in the initial
batch and lower in the comparative-summary batch.
In one Scenario A pair in the comparative-summary batch, the summary condition
finds a best AUC 0.017284 higher than the raw-evidence condition; the other two
pairs show a small gain and a decline. In Scenario E, two pairs
reach the same final best AUC, but the summary condition arrives eight
selections earlier in one pair and eight later in the other; the third pair
ends lower. Across both batches, summaries do not consistently improve
best AUC or reach the same result sooner.

\section{Related Work}
\label{sec:related-work}

\paragraph{Agentic research for recommendation.}

AutoRecLab~\cite{autoreclab} makes requirements explicit and uses prototype
execution and refinement to implement studies with recommendation libraries.
Industrial model changes also need to preserve the intended mechanism and
satisfy production constraints. NOVA~\cite{nova} addresses this through semantic
verification, candidate testing, and trajectory memory that guides subsequent
architecture changes, including adaptations from research papers.
RecHarness~\cite{recharness} studies how to allocate trials across modification
directions: a bandit selects a direction, while an LLM constructs the concrete
hypothesis and code change. CORAL~\cite{coral} applies a related feedback loop
to a different optimization surface, adjusting retrieval and serving
configurations under operating constraints using measured online outcomes.
AutoLR~\cite{autolr} combines proposal review, evidence-guided direction
selection, and offline evaluation to prepare candidates for human-gated
online testing and launch review.
RecSys Factory concentrates agent autonomy at decision points within
deterministic industrial pipelines~\cite{recsysfactory}.
Our setting concerns model research within business-defined input and output
interfaces, with architecture and training changes evaluated before deployment.

Within this setting, our earlier AgentX work~\cite{agentx} established
paper-driven exploration, multi-round experimentation, and cross-paper
combination. Work on token mixing and feature interaction, including
RankMixer~\cite{rankmixer}, Climber~\cite{climber}, and
UniMixer~\cite{unimixer}, provides architectural mechanisms for such
exploration. From Trajectories to Evidence~\cite{trajectories} examined how to
turn experimental histories into records supported by code, measurements, and
applicability conditions, and studied their conditional use in subsequent
adaptations. This report builds on that foundation to examine sustained
research paths. Returned implementations and findings inform the next
proposal's question, starting code, and evaluation target. We follow how
those choices support continued improvement, combination, or diagnosis when
new observations change the research objective.

\paragraph{Long-horizon research and experience reuse.}

Long-horizon research systems retain model findings and execution experience
across experiments~\cite{longhorizonengineering,scienceflow}.
Auto-RecSys~\cite{autorecsys}
supports this reuse in industry-scale recommendation through asynchronous
experimentation, shared memory, and two feedback loops. Its execution loop
updates model-specific playbooks from operational experience; its idea loop
uses experimental outcomes to guide subsequent proposals. Ideas can come
from researchers, papers, and model-aware brainstorming, and experiment
history helps filter duplicates and develop combinations of partial successes.
This is closely related to our separation between forming research questions
and conducting the resulting experiments, as well as to the Model Agent's
reuse of implementation experience.

Evaluated implementations also support cumulative search beyond recommendation.
AlphaEvolve~\cite{alphaevolve} combines LLM-generated program changes with
evaluation and evolutionary selection. ERA~\cite{era} uses tree search to
develop empirical software, incorporates external research ideas, and explores
recombinations of promising methods. Agora~\cite{agora} instead emphasizes
shared research state: Git-backed contributions preserve dependencies and
verification records across agents, with a sustained weight-transfer study
illustrating their use. These approaches, together with memory-guided graph
search~\cite{mlevolve}, make earlier solutions and findings material for further
investigation. We examine the corresponding choices in
industrial model research: which measured round to inherit, whether a
combination improves on its strongest source, and how business feedback can
redirect a successful model toward calibration or other deployment
requirements. Binding these choices to code and evaluation context allows
later tasks to continue from a specific result instead of reconstructing it
from a narrative summary.

\paragraph{Research selection and historical replay.}

When several continuations are available, experience reuse leaves a further
decision: which opportunity should receive the next experiment budget?
AIRA-dojo~\cite{airadojo} separates search policies from the operators that
generate or revise solutions, comparing greedy, Monte Carlo tree search, and
evolutionary strategies. AI Research Preference Models~\cite{rpm} address the
cost of candidate evaluation by using plans, code, and previous outcomes to
predict which candidates merit execution. Dream-RSI~\cite{dreamrsi} instead
uses recorded discovery trees as replay environments for evaluating and
revising exploration policies before deploying them in further online search.
These works separate solution quality from the policy that allocates research.

Our benchmark studies this distinction at the level of complete research
experiments. A selected experiment may itself contain multiple implementation and
evaluation rounds; those rounds become evidence for later selections.
The replay benchmark compares how policies allocate opportunities among
Reproduce, Follow-up, and Composition while respecting their recorded
dependencies. Fixed task-type rotation provides the baseline, with the agent
selecting a candidate within the allocated type; bandit and joint selection
vary that allocation. Replay evaluates these choices on observed research
paths, while the production cases show what continuing those paths entails:
retaining useful implementations, testing combinations, and resolving
questions raised by prior results.

\section{Lessons from Long-Horizon Model Research}
\label{sec:discussion}

Our observations suggest six lessons, grouped around
three themes: \emph{preserve the state needed to continue research} (L1--L3),
\emph{let evidence refine the research question} (L4), and
\emph{match research decisions to their evidence and dependencies} (L5--L6).

\paragraph{L1. Preserve the best measured implementation alongside the latest research state.}
Retain the best measured code and its measurements even when research moves
on to a later version. Many multi-round experiments finish below an earlier
measured best (Section~\ref{sec:continuation-results}), so overwriting it can
remove a useful starting point. Keep the latest diagnosis and failed attempts
alongside it: they explain what has been tried since that model was obtained.

\paragraph{L2. Distinguish parent-level AUC gains from a new path-best result.}
Specify the starting code separately from the results used to judge progress.
In Scenario A, continuations more often record AUC above their direct parents
than above all comparable ancestors (Section~\ref{sec:continuation-results}).
Using both references distinguishes recovery from a new path-best record.
The strongest ancestor need not supply the starting code: a repair may need
the diagnostic implementation and its observations.

\paragraph{L3. Carry evaluation context together with implementations.}
Retain the prediction output, label, sample population, metric, and comparison
result with the code. These details determine what a continuation is testing.
In the calibration case (Section~\ref{sec:pcoc}), review caught a draft
targeting a different prediction task (Appendix~\ref{app:auditor-cases}).
Reviewing the proposed change together with its evaluation helps keep the
next experiment focused on the inherited observation.

\paragraph{L4. Let business feedback update the research objective.}
A continuation should specify both the new requirement and the performance
to preserve. In the calibration path, business feedback redirected research
on a model with ranking gains toward prediction bias (Section~\ref{sec:pcoc}).
That changed the purpose of the next experiment: diagnosis needed to
distinguish explanations, while repair needed to improve calibration and
check the retained ranking performance.

\paragraph{L5. Treat candidate availability as part of experiment selection.}
In the Scenario A replay, agents did not select all the source experiments
needed for Composition, so preferring that action type alone did not make
its candidates available (Section~\ref{sec:rq4}). A next test should compare
policies using only an experiment's expected result with policies that also
value the combinations it unlocks.

\paragraph{L6. Evaluate experience reuse according to the decision it supports.}
Transferred experience supplied new proposals implemented and evaluated in Scenario D
(Section~\ref{sec:knowledge-transfer-results}), while adding summaries to raw
evidence did not consistently improve best AUC or shorten selection paths
(Section~\ref{sec:rq4}). Instruction revisions improved another output:
first-draft implementation plans (Appendix~\ref{app:model-evolution-results}).
These are three distinct uses of experience. New model ideas need evaluation
in the target setting; implementation guidance needs checks on newly generated
outputs; selection guidance needs comparison of the choices, models, and path
costs it produces.

Across these investigations, AgentX-Model has produced and evaluated a variety
of model modifications. We present selected methods in
Appendix~\ref{app:unexpected-findings}; the main text focuses on how model
changes arise from prior experiments and how their results support further
research.

\section{Conclusion}
\label{sec:conclusion}

This report examines how AgentX-Model carries the results of one experiment
into subsequent model research. Production records document sustained operation
and offline model improvements within business-defined sandboxes. Online A/B
tests recorded gains in acquisition efficiency, target-segment advertising
spend, and engagement, with task pruning also reducing model cost.

The research paths show what continuing this work entails: retaining code
from useful measured rounds, choosing references that distinguish inherited
performance from new progress, and pursuing questions raised by the results.
Business feedback extended one such path from ranking gains to diagnosis and
offline calibration improvement. When choosing among further experiments,
fixed task-type rotation with agent-based candidate selection remained
competitive in historical replay. The combinations left unavailable by missing
source experiments point to a next question for selection: how to pursue the
experiments that make useful continuations possible.

\clearpage
\appendix
\part*{\centering Appendices}
\addcontentsline{toc}{part}{Appendices}

\section{Research Agent Proposal Refinement}
\label{app:proposal-refinement}

\subsection{Proposal Refinement Mechanism}

In one calibration proposal, the draft targeted a different prediction task
from the problem it was intended to repair. In a Composition proposal, review
found that the design repeated one source instead of retaining a distinct
contribution from each. These cases, detailed in
Appendix~\ref{app:auditor-cases}, illustrate why the Research Agent reviews a
proposal before committing training resources to it.

\begin{figure}[!htbp]
\centering
\includegraphics[width=\textwidth]{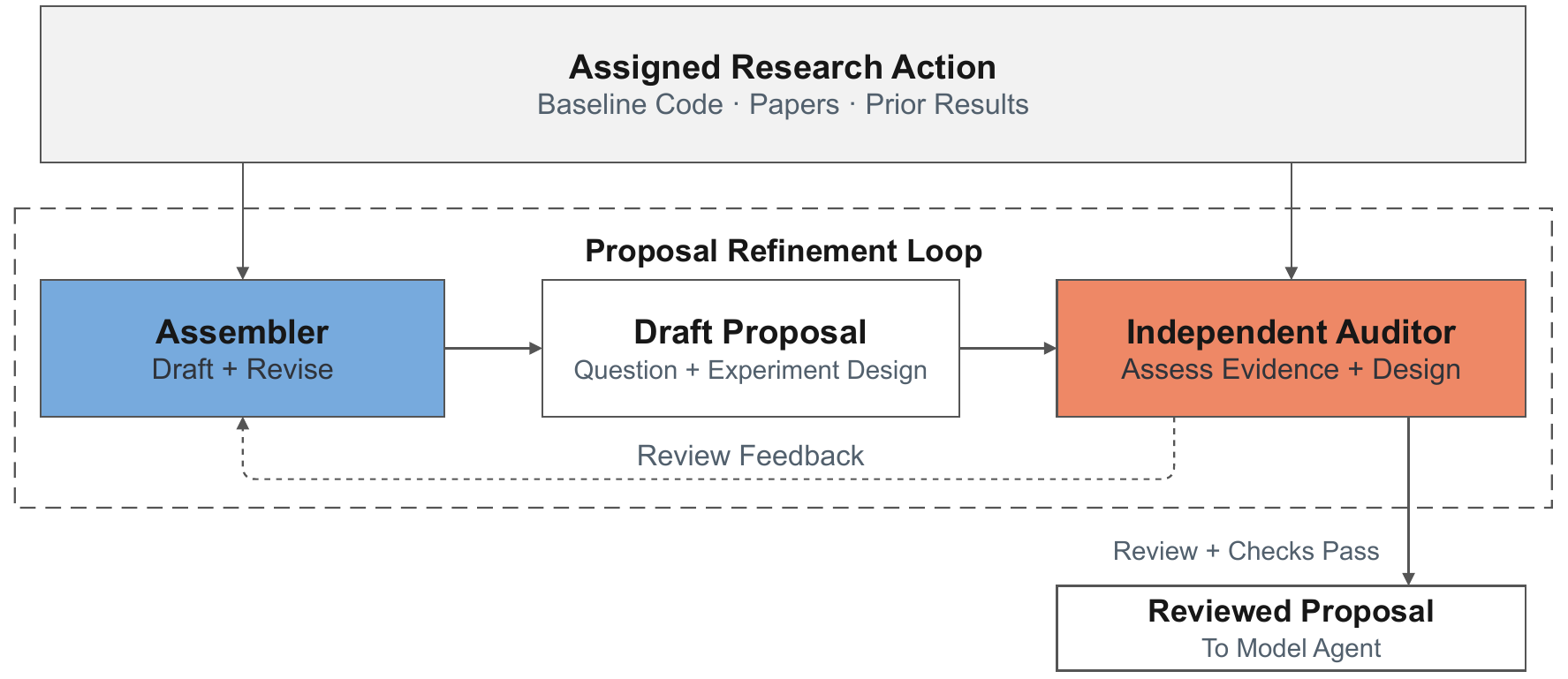}
\caption{Proposal refinement inside the Research Agent. The dashed enclosure
marks the core loop: construct a proposal, review it independently, and revise
and resubmit in response to feedback. Assigned actions and shared evidence are
inputs; delivery to the Model Agent is an exit after approval and automated
checks. Execution limits are enforced by the harness.}
\label{fig:research-agent-refinement}
\end{figure}

The review loop in Figure~\ref{fig:research-agent-refinement} separates
construction from assessment. An Assembler investigates the assigned research
action and develops the proposal. An Auditor reviews it in a separate context
and returns the issues that must be addressed before delivery.

\paragraph{Independent review with a shared understanding of the baseline.}
Both roles receive the baseline's inputs, prediction tasks, model structure,
and supporting code references. They reuse a checked description of that
version when available, or establish it by inspecting the code. They share
these facts but retain separate responsibilities: the Assembler writes and
revises the proposal, and the Auditor decides whether it is ready for delivery.

The Auditor uses supporting evidence, expected observations, and unresolved
assumptions to assess whether the change
$\delta_t$ to starting implementation $s_t$, together with the evaluation
$v_t$, can answer $q_t$. For a Composition, it examines what each source
contributes after overlapping changes are removed. For a Diagnose, it asks
which observation could contradict the
proposed explanation.

\paragraph{Focused revision after review.}
The Assembler addresses the most important issue in the review and resubmits
the proposal. It retains established baseline facts and reads further when
a revision reaches a new code path or encounters conflicting evidence.
The Auditor then checks whether the revision resolves the issue. If the
candidate remains unsupported, the agent can try a recorded alternative or
defer. Time, revision, and no-progress limits halt attempts that cannot secure
approval.

\paragraph{Approval of the reviewed proposal.}
Delivery requires both the Auditor's approval and automated checks for required
information, source references, and matching model versions. The two checks
serve different purposes: a code reference may exist yet point to an operation
unrelated to the proposed change. Only the exact reviewed version is delivered;
later edits require another review. Drafts, review feedback, and revisions are
retained for later analysis of design changes.

The following cases show how review changed delivered proposals;
Section~\ref{sec:rq1} reports end-to-end proposal-production time.

\subsection{Proposal Revision Cases}
\label{app:auditor-cases}

\paragraph{Correcting the target of calibration repair.}
The PCOC Follow-up in Section~\ref{sec:pcoc} initially proposed calibrating
a different prediction output. The Auditor checked the parent's report and
code and identified the mismatch with the reported bias.
It also found that the draft cited raw
business-baseline code without establishing the required starting point in
the parent's SMES implementation. The Assembler revised the question,
code-change instructions, and evaluation together. The delivered proposal
targeted the original prediction output and label, and retained
the parent's SMES routing and training-side calibration. It required paired
calibration measurements and a check that ranking performance was preserved.
One blocking review led to one revised submission;
the interval from the first review request to delivery was approximately
20 minutes.

\paragraph{Correcting a composition that repeated one source.}
In another Scenario A proposal, the Assembler attributed an attention output
projection, $W_O$, to the second parent and proposed adding it to the first.
The Auditor inspected both parents' code changes and found that the first
already contained $W_O$: the proposed combination repeated that parent and
omitted the second parent's layer normalization of the reverse-process
representation. The Assembler rewrote the question to test that normalization
on the first parent's implementation, taking into account its previously
observed lack of AUC improvement in the second parent's configuration.
A second review found that the revision placed normalization before the wrong
expert-fusion component and checked mechanism activation without requiring
the AUC improvement posed by the question. The final proposal corrected the
insertion point and added an AUC-based completion condition. Two blocking
reviews led to two revised submissions, with approximately 27 minutes from
the first review request to delivery.

In both cases, the first submitted draft had passed automated validation.
The Auditor's review changed the research design by checking it against the
source experiments.

\subsection{Workflow Evolution and Development Observations}
\label{app:research-agent-evolution}

Our Research Agent initially used language models within an engineer-defined
sequence of reading, checking, and repair. New paper mechanisms and business
baselines often required another coded branch to decide what to read or repair.

\begin{figure}[!t]
\centering
\includegraphics[width=\textwidth]{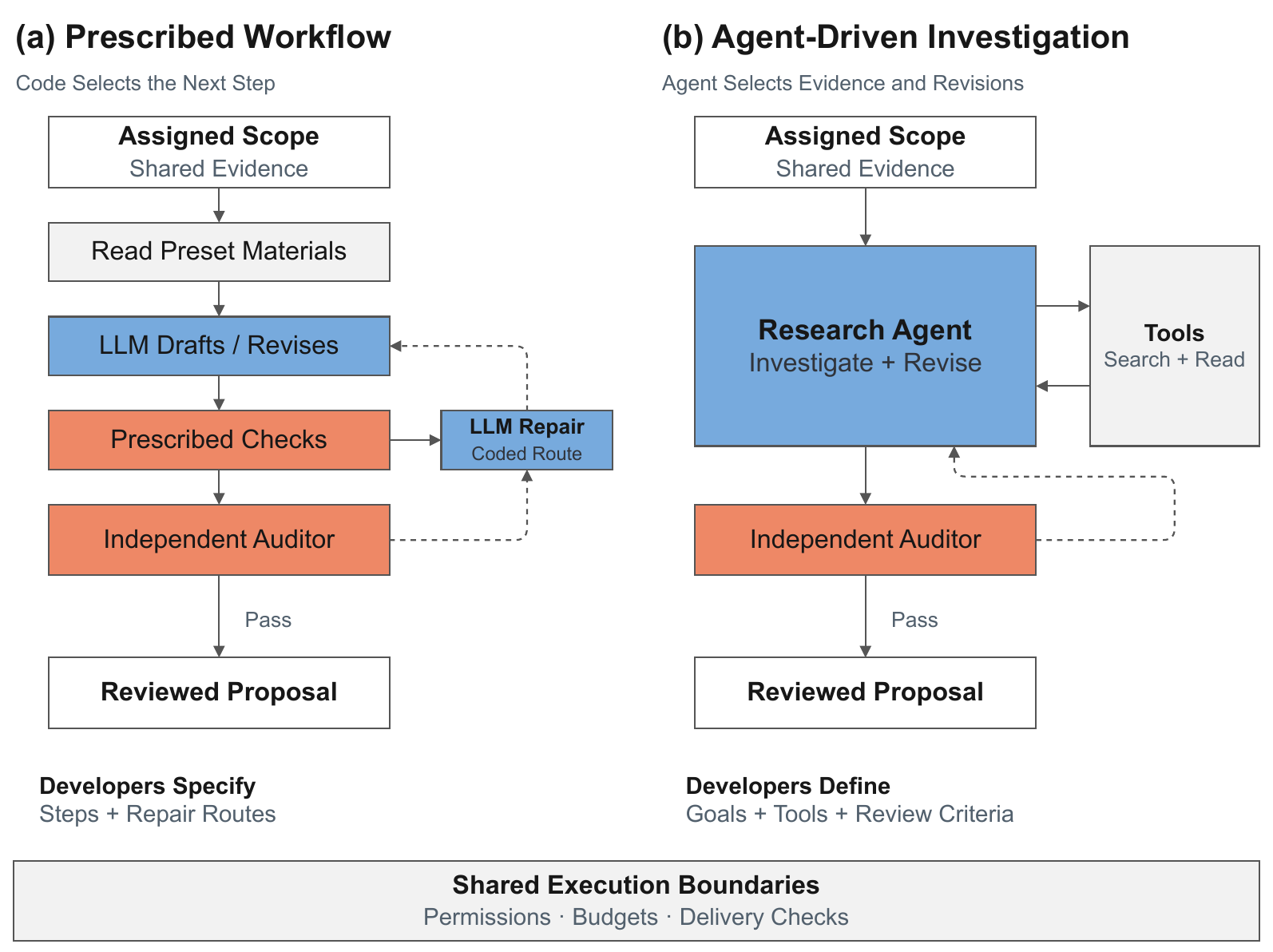}
\caption{From prescribed workflows to agent-driven proposal production.
Both versions use language models and the same independent Auditor; the
investigation and repair path is prescribed by code in (a) and selected by the
agent in (b). Dashed arrows indicate revision feedback. Execution boundaries
remain enforced in both versions.}
\label{fig:proposal-evolution}
\end{figure}

We moved these decisions into the Research Agent's loop
(Figure~\ref{fig:proposal-evolution}). Within the allocated task type, it
chooses evidence to inspect and revises its proposal in response to the same
independent Auditor. The harness retains control over permissions, budgets,
output checks, and delivery.

Development consequently shifted toward evidence-access tools, context
organization, review feedback, stopping conditions, and delivery.
The agent could keep reading without committing to a candidate, carry excess
context, or reopen a broad investigation after a small review comment. We
therefore separated exploration from revision: first submit a complete
proposal, then address the issue preventing its approval
(Section~\ref{sec:refinement}).

\section{Model Agent Execution and Self-Evolution}
\label{app:model-agent}

\subsection{Multi-Round Model Research}

\begin{figure}[!ht]
\centering
\includegraphics[width=\textwidth]{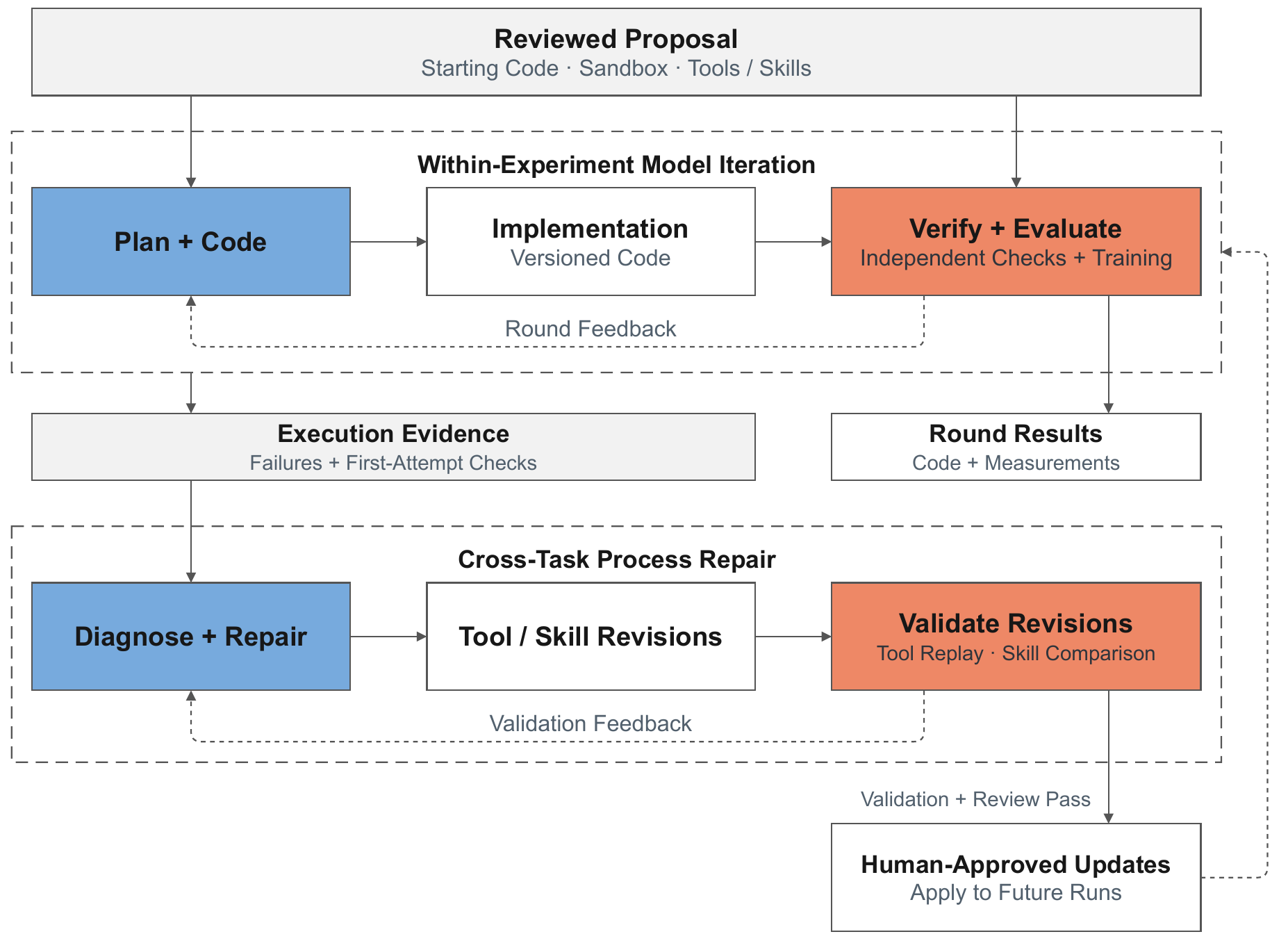}
\caption{Model Agent iteration and process repair. The upper loop develops a
model within the reviewed experiment. Failures and first-attempt check results feed
the lower repair loop, where tool and skill revisions undergo separate forms
of validation before human-approved deployment to future runs. Repair acceptance
rules remain fixed. Cross-task experience learning is described separately.}
\label{fig:model-agent-loops}
\end{figure}

The Model Agent investigates the reviewed question through repeated cycles of
planning, coding, verification, training, and evaluation. Starting from $s_t$,
it can revise the model or test alternatives within the proposal's scope,
using $v_t$ to assess the results. A weak result may motivate a different
implementation; training instability may require additional observations
before another model change. Figure~\ref{fig:model-agent-loops} shows this within-experiment loop and how
its execution records support later improvements to the agent itself.

\paragraph{Checking the implementation before training.}
Each round begins with a plan linking the intended model change to its rationale
and observable tests. Structural checks verify the required elements and code
references, while a language-model review checks fidelity to the research
materials. After coding and self-checking, the implementation undergoes a smoke
test and independent verification against the plan. The verification decision
is also checked against its evidence: cited code must exist, claimed observables
must be implemented, and the conclusion must follow from those checks.

Findings return to the relevant generation stage for bounded revision. A round
that exhausts its repair budget retains a failure record rather than advancing
to training. First-attempt check results remain available after successful
repair to track recurring defects.

\paragraph{Using experiment feedback.}
After training, the Model Agent compares results with the reference
implementation and checks its explanation against the recorded observations.
These findings, unresolved questions, and human comments guide the next round.
Each round's code and measurements are retained for later inheritance,
including useful intermediate models (Section~\ref{sec:evidence}).

\subsection{Self-Evolution Through Process Repair}
\label{app:model-improvement}

Correcting an implementation helps the current experiment. When the same
problem recurs across experiments, the reusable tool or instruction may need
to change. Process repair investigates this second question: what change to
the execution process would prevent later tasks from needing the same local
correction? The language model's weights remain fixed; the objects of revision
are error classifiers, checkers, prompt construction, and skill instructions.
Memory provides a separate way to carry implementation experience into later
tasks (Appendix~\ref{app:model-memory}).

\paragraph{Turning repeated failures into a repair target.}
The harness groups failure records and first-draft check findings using
deterministically extracted signatures: the recorded reason, exception,
gateway error, and checker finding. This gives recurring failures a stable
identity across experiments. The repair agent inspects representative
records and the deployed code, identifies the component responsible, and
proposes a bounded change in an isolated workspace. It can revise permitted
components but cannot change the evidence or its own acceptance rules.

The component matters because similar execution outcomes require different
repairs. A missing native library can terminate training just as a model-code
bug does, but the recovery actions differ. In the documented \texttt{libjvm}
case, the repair changes the failure classifier from a code-error decision
to an environment-error decision eligible for retry. The patch addresses
the system's response to the failure; repairing the external environment
remains a separate operation. For repeated omissions in an implementation
plan, the relevant target is instead the instruction that generates the plan.

\paragraph{Replaying a tool change against fixed evidence.}
A classifier or checker patch can be tested on the same archived inputs
before and after the change. The validation set contains the target failures
and successful controls with known expected outcomes. It measures whether
the target decisions are corrected and whether previously correct decisions
are preserved. In the missing-library case, validation checks that environment
failures are classified correctly and become eligible for retry.

Independent review examines the patch and its diagnostic rationale. Replay
checks decisions against expected outcomes while holding the inputs and
sample membership fixed.

\paragraph{Regenerating outputs after an instruction change.}
Rechecking an existing plan cannot measure whether a new instruction produces
a better plan. Skill validation instead generates fresh outputs with both
the current and candidate instructions. The automated repair protocol separates
cases available during patch development from held-out target and control
cases used for acceptance. Repeated current-version runs estimate variation;
the comparison checks the targeted defect and other errors in newly generated
outputs.

Version isolation is part of this test. Each instruction version is loaded
in an isolated execution environment outside the working repository, with a
distinct marker used to check which version was read. The repair is limited
to the diagnosed instruction section. Checks examine both the declared
categories and whether required observations appear in the generated content.

\paragraph{Deploying and observing the revision.}
The test split and acceptance rules remain outside the repair agent's control.
Validated revisions are submitted for human approval before production use.
After deployment, subsequent execution records are checked for recurrence
of the same failure signature; recurring failures can reopen the investigation.
Appendix~\ref{app:model-evolution-results} evaluates an earlier instruction
revision using fresh plans for a fixed input batch.

\subsection{Agent-Initiated Memory Extraction and Reuse}
\label{app:model-memory}

Some findings are better retained as implementation guidance than encoded
as changes to the tools. The Model Agent autonomously assesses whether an
experiment has produced reusable experience and, when it has, extracts a
candidate memory for human review. Approved memories enter production to
guide subsequent planning and coding within their model scope. Each candidate
records the source tasks, supporting code and measurements, and conditions
under which the guidance applies. It may describe a recurring implementation
mistake or a practice supported by a measured improvement.

\paragraph{Using a memory in a later implementation.}
For a later task, the Model Agent examines the memory's source code,
measurements, and conditions to decide whether the guidance applies to the
current model. It uses applicable guidance in planning and coding, then
evaluates the resulting implementation. Memory guides individual tasks;
process repair updates shared tools and instructions.

\subsection{Instruction Revision Evaluation}
\label{app:model-evolution-results}

Repeated defects in implementation plans can be addressed by improving the
instructions used to generate later plans. In a fixed-batch regression evaluation, we
compare the original and revised instructions on the same 40 paper-reproduction
inputs whose earlier failures informed the revision. The Model Agent proposed
changes to its plan-generation skill and the constraint checklist supplied during plan
repair; a human reviewed and approved the proposed changes. These are shared
instructions used to generate plans across tasks. Each of the 40 inputs is
used to generate one new plan under each instruction version.

Table~\ref{tab:model-agent-evolution} compares first drafts generated before
and after the revision. Both versions use the same evaluation procedure:
deterministic checks assess the required plan elements, while an LLM judge
evaluates fidelity to the mechanisms specified in the input material.
We also count cases that omit a required
mechanism without acknowledging the omission.

\begin{table}[htbp]
\centering
\small
\setlength{\tabcolsep}{9pt}
\caption{First-draft implementation plans before and after instruction revision
in a regression evaluation on the same 40 paper-reproduction inputs.}
\label{tab:model-agent-evolution}
\begin{tabular}{@{}lrr@{}}
\toprule
Metric & Before & After \\
\midrule
First-draft structural pass rate & 37.5\% & \textbf{95.0\%} \\
Semantic fidelity & 72.5\% & \textbf{95.0\%} \\
Plans with unacknowledged mechanism omissions & 4 & \textbf{0} \\
\bottomrule
\end{tabular}
\end{table}

Structural pass rate increased by 57.5 percentage points and semantic fidelity
by 22.5 points, with no remaining unacknowledged omissions in this batch.
The revised instructions improved first drafts before any per-plan repair.

\clearpage
\section{Selected Findings from AgentX-Model Research}
\label{app:unexpected-findings}

The four cases adapt published methods to weight features using within-sample
relationships, pair continuous features with discrete prototypes, generate
missing semantic profiles, and revise code selection and updates in a quantizer.

\subsection{Mem-GF: Parameter-Free, Sample-Adaptive Feature Weighting}
\label{app:memgf-adaptation}

\begin{figure}[!htbp]
\centering
\includegraphics[width=\textwidth]{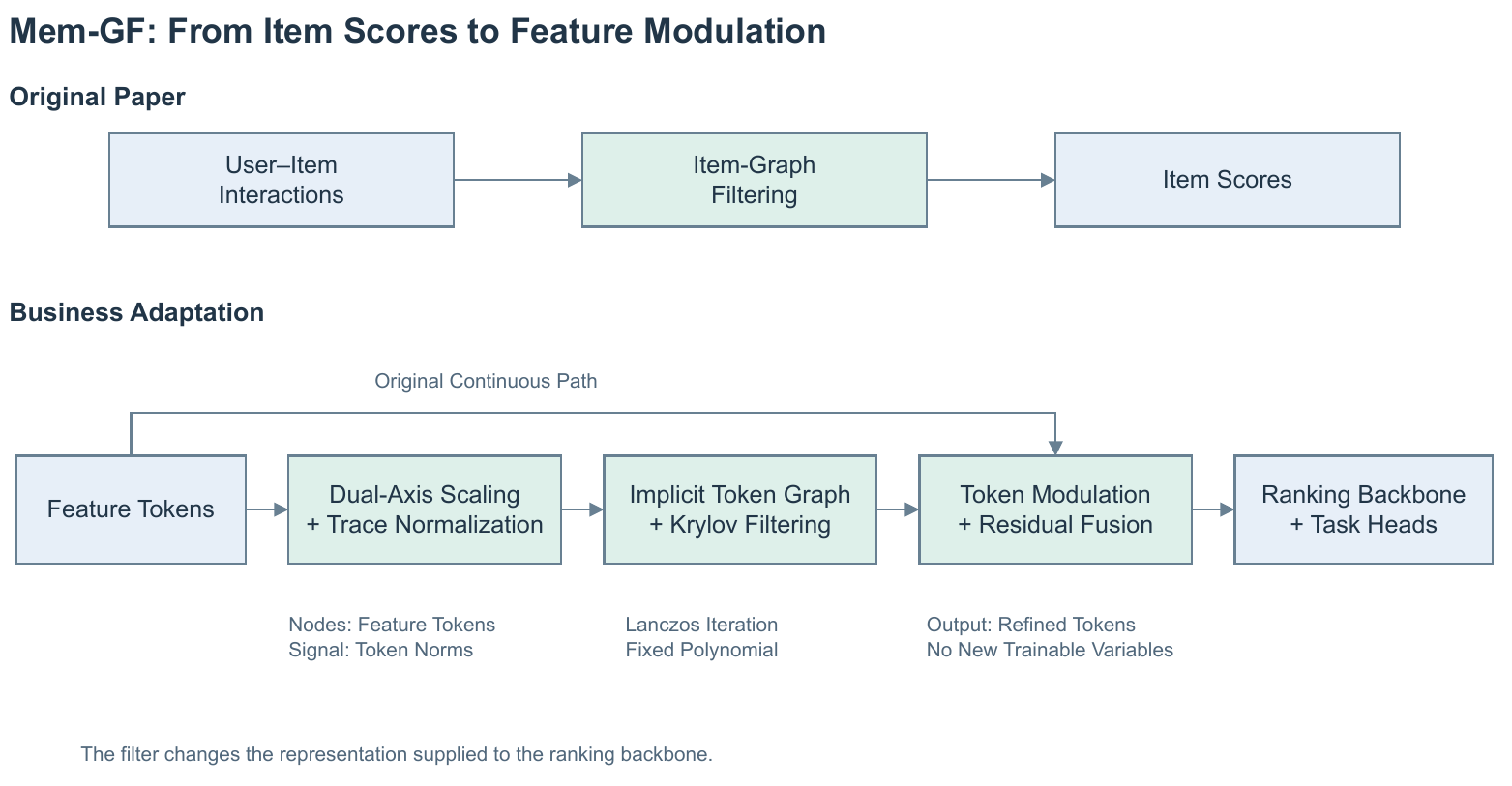}
\caption{Sample-adaptive feature modulation. Within-sample token relationships
determine the adjustments supplied to the ranking backbone.}
\label{fig:memgf-adaptation}
\end{figure}

This adaptation dynamically adjusts features using relationships within the
current sample, without learning an additional weighting network.
Mem-GF~\cite{memgf} originally filters user interaction signals on an item
graph, using Krylov/Lanczos computations to apply a polynomial filter without
storing the full item similarity matrix. The adaptation instead treats a sample's feature tokens as graph
nodes, using their relationships to adjust the ranker inputs
(Figure~\ref{fig:memgf-adaptation}).

The branch rescales and normalizes the tokens. Their inner products define an
implicit similarity operator, applied through matrix--vector products rather
than an explicitly stored graph. Krylov/Lanczos computations apply the polynomial
filter to a signal formed from the token norms. The resulting scalar
$s_i$ modulates token $i$ before residual fusion:
\begin{equation}
H'_i=\operatorname{L2Norm}\bigl(0.9H_i+0.1s_i\bar H_i\bigr),
\label{eq:memgf-fusion}
\end{equation}
where $H_i$ is the original token and $\bar H_i$ is its rescaled version.
Both the weighting and coordinate rescaling depend on the sample, allowing
the branch to change the direction of its feature representations. The
added branch has no trainable parameters. In Scenario E, AUC increased from the business
baseline's \textbf{0.815403 to 0.820512}, a gain of 0.005109.

\subsection{TokenMinds-Inspired Quantization: Pairing Features with Shared Prototypes}
\label{app:tokenminds-adaptation}

\begin{figure}[!htbp]
\centering
\includegraphics[width=\textwidth]{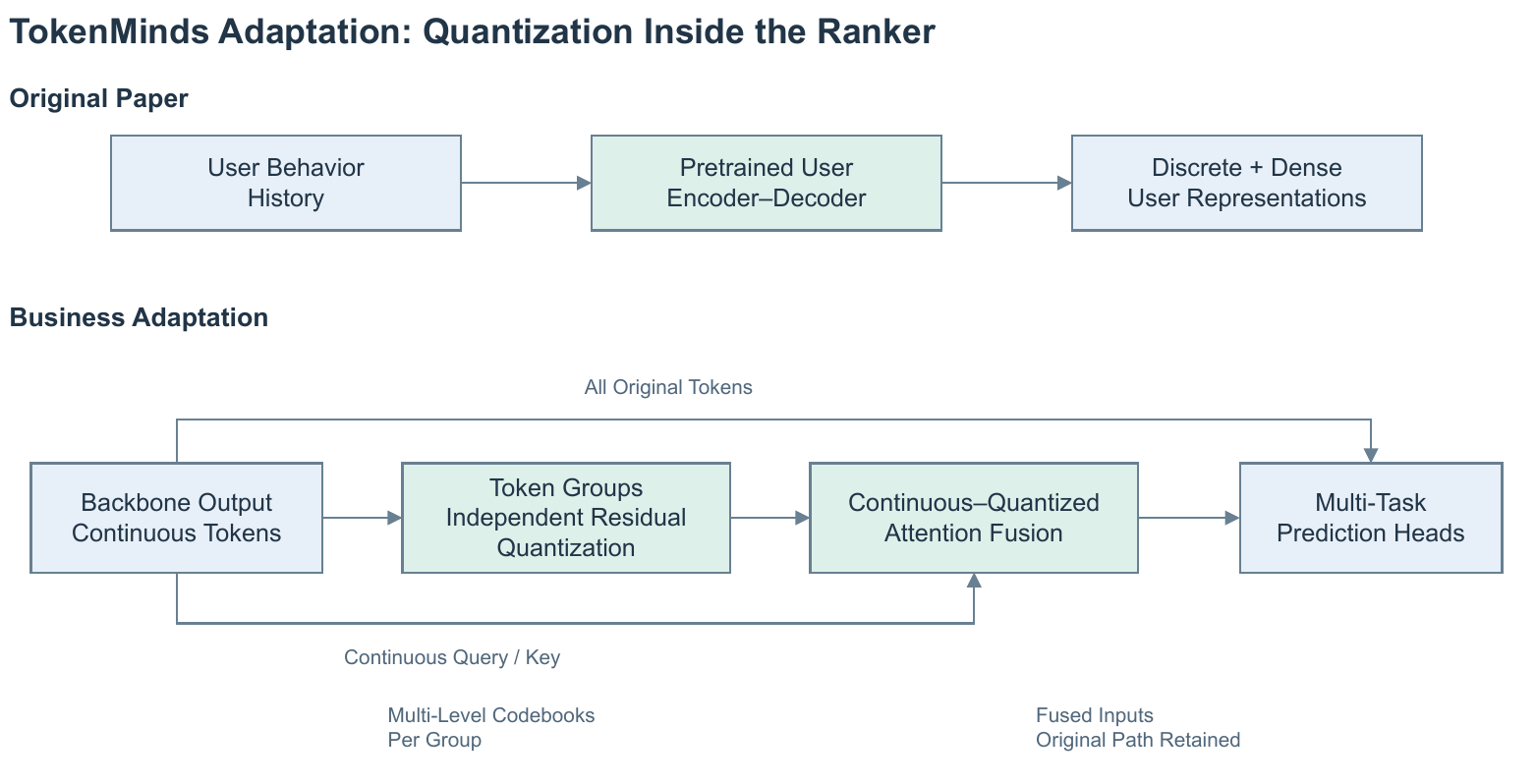}
\caption{Continuous features read learned discrete prototypes. The fused
representations and all original tokens feed the prediction network.}
\label{fig:tokenminds-adaptation}
\end{figure}

Continuous intermediate representations preserve sample-specific detail;
learned discrete prototypes offer shared representation patterns.
TokenMinds~\cite{tokenminds} constructs semantic IDs from video content, then
uses a pretrained user model to produce discrete user tokens and continuous
embeddings for downstream models. This adaptation borrows their complementary
roles: the ranking network directly quantizes its intermediate tokens into
learned prototype vectors while retaining the original continuous features
(Figure~\ref{fig:tokenminds-adaptation}).

Token groups use separate multi-level residual quantizers. Each level
selects a code vector, and the next represents what remains after that
selection; the selected vectors are added to form the quantized representation.
Fusion uses continuous features as queries and keys and quantized features
as values, letting the continuous representation determine how to read the
prototypes. The fused vectors join all original tokens in the multi-task
prediction network. The prototypes are learned vectors, not predefined
user categories.

This implementation originated from a human-specified research request and
is already part of the Scenario D baseline.

\subsection{LGCD with AFTM: Generating Missing Semantic Profiles}
\label{app:lgcd-adaptation}

\begin{figure}[!htbp]
\centering
\includegraphics[width=\textwidth]{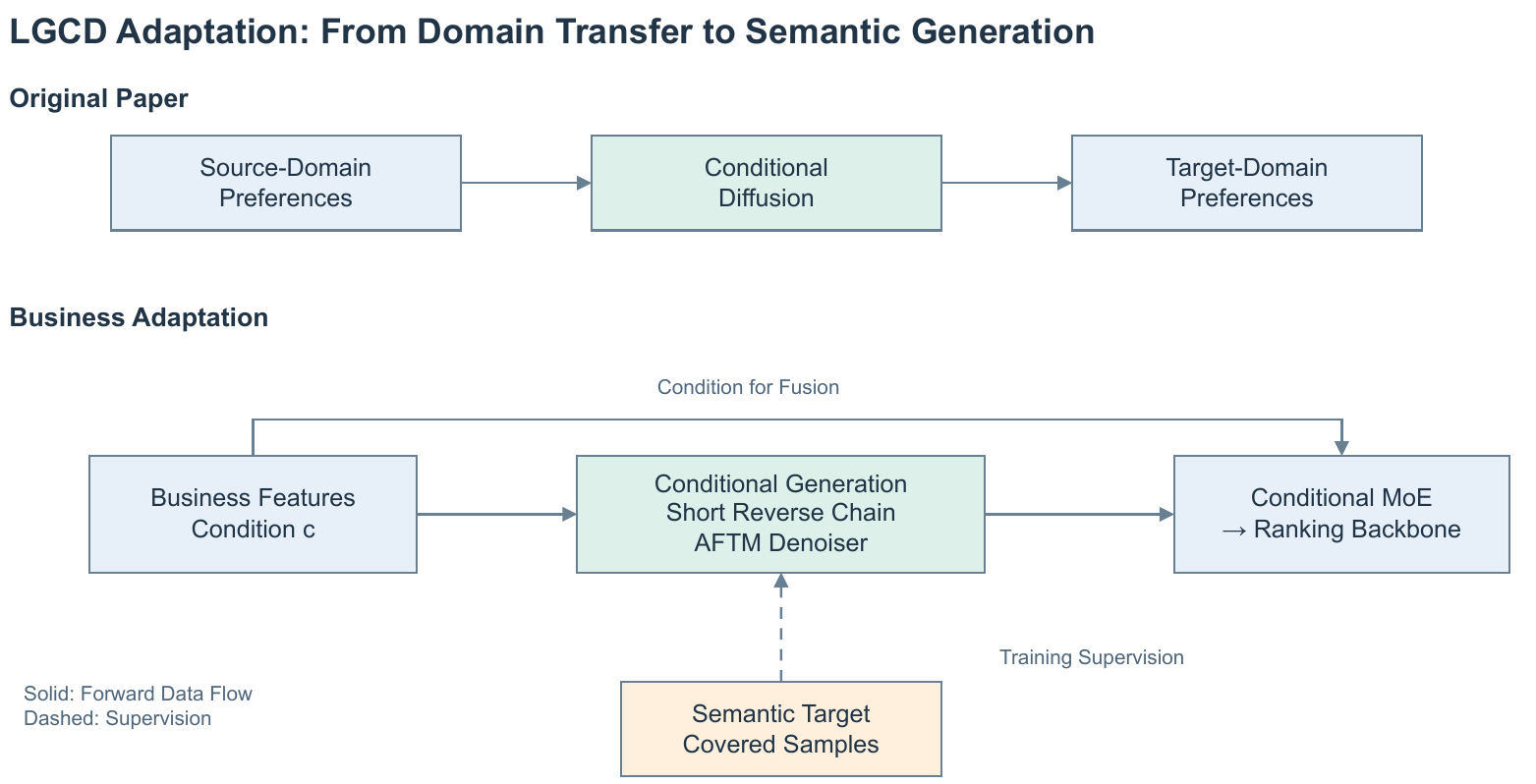}
\caption{Generating semantic profiles from business features. Existing profiles
supervise training; generated representations serve all samples.}
\label{fig:lgcd-adaptation}
\end{figure}

LGCD~\cite{lgcd} uses diffusion to generate a user's preferences in another
domain. The Scenario A adaptation instead generates a semantic vector from
business features, using existing semantic embeddings as training targets.
The ranker uses the generated vectors for all users, including those without
an existing semantic profile (Figure~\ref{fig:lgcd-adaptation}).

Training pairs business user features with available semantic embeddings.
The embeddings are split into blocks and normalized to form denoising
targets. A conditional denoiser learns to reconstruct them, while an
initial-state predictor learns to generate a starting representation from
the business features. The ranking path uses this predictor followed by a
short reverse diffusion chain and fuses the generated representation back
into the ranker. Reconstruction trains the denoiser; the reverse chain is
detached from the ranking-loss gradient. For users without target embeddings,
the initial-state predictor also learns to match the detached generated output.

LGCD replaced the original vector-quantization path, removing the insertion
point used by the AFTM parent. The Composition therefore moves AFTM-inspired
mixing into the denoiser. In FuXi-$\beta$, AFTM~\cite{fuxibeta} mixes sequence
content using positional and temporal relations, then applies multiplicative
gating. Here, there is only one condition token, so standard cross-attention
has a softmax weight of one along the condition axis. The adapted gate uses
the noisy state and diffusion time to modulate each dimension of the user
condition, adapting sequence mixing to state-dependent conditioning.

The complete implementation reached \textbf{0.779965}, compared with the business baseline
of 0.773683 and the strongest direct parent of 0.778906. The corresponding
recorded margins are 0.006282 and 0.001059; Figure~\ref{fig:composition-case} shows the
associated experimental trajectory.

\subsection{ORQ with DIGER: Changing How a Quantizer Selects and Updates Prototypes}
\label{app:orq-diger-adaptation}

This Composition keeps an existing quantizer's progressive compression
structure but changes how it selects and updates code vectors.
Orthogonal residual quantization (ORQ) from DOS~\cite{dos} first rotates
features, selects dimensions for quantization, and passes the remaining
dimensions and quantization residual to the next level. Concentrated code
usage in the Scenario A parent motivated the Research Agent to insert
DIGER-inspired exploration~\cite{diger} into the existing code-assignment steps.

During training, historical usage identifies frequently selected codes,
whose assignment scores receive Gumbel noise to encourage exploration.
The forward pass selects one code vector, while gradients follow a soft
assignment. Soft assignment statistics also contribute to exponential-moving-
average codebook updates. Inference removes the exploration noise and selects
one code vector. ORQ's rotation and residual propagation remain;
dimension selection uses a hard mask that does not pass gradients to its
scoring network through that operation.

The best complete implementation, including numerical protection of its
auxiliary loss, reached \textbf{0.778186}. The ORQ and DIGER parents reached
0.775863 and 0.775681, respectively: the combination improves on the stronger
direct parent by 0.002323 and on the business baseline of 0.773683 by
0.004503. Whereas the TokenMinds-inspired case adds a complementary
representation, this case changes selection and learning inside an existing
quantizer.

\end{document}